\documentclass[copyright,creativecommons,noncommercial]{eptcs}  
\providecommand{\event}{SEMSP}

\usepackage{graphicx}    
\usepackage{times}
\usepackage{tabulary} 
\usepackage{array}  

\usepackage{tikzfig}
\usepackage{stmaryrd}    
\usepackage{docmute}    
\usepackage{keycommand}    

\usepackage{rotating}
\usepackage{floatpag}
\usepackage{xifthen}

\usepackage{bm}

\rotfloatpagestyle{empty}

\newcommand{\warningsign}{\raisebox{1pt}{\fontencoding{U}\fontfamily{futs}\selectfont\char 66\relax}\xspace}

\ifthenelse{\isundefined{\showoptional}}{
  \newcommand{\showoptional}{1}
}{}

\ifthenelse{\isundefined{\ismain}}{
  \newcommand{\ismain}{0}
}{}

\ifthenelse{\isundefined{\lecturenotes}}{
  \newcommand{\lecturenotes}{0}
}{}

\ifthenelse{\isundefined{\cupversion}}{
  \newcommand{\cupversion}{0}
}{}

\usepackage{hyperref}

\usepackage{amsmath,amsthm,amssymb}
\usepackage{wasysym}

\usepackage{xspace,enumerate,color,epsfig} 
\usepackage{graphicx}
\graphicspath{{.}{./figures/}}

\usepackage{tikzfig}
\usepackage{stmaryrd}
\usepackage{docmute}
\usepackage{keycommand}

\newcommand{\xhalf}[1]{\,p^{\textrm{$\frac{1}{2}$\!\!}}}
\newcommand{\xhalfinv}[1]{\,p^{\textrm{-$\frac{1}{2}$\!\!}}}

\newkeycommand{\boxmap}[style=small box][1]{\,\tikz{\node[style=\commandkey{style}] (x) {$#1$};\draw(0,-1.25)--(x)--(0,1.25);}\,}
\newkeycommand{\smallboxmap}[style=small box][1]{\,\tikz{\node[style=\commandkey{style}] (x) {$#1$};\draw(0,-1.25)--(x)--(0,1.25);}\,}
\newkeycommand{\shortboxmap}[style=small box][1]{\,\tikz{\node[style=\commandkey{style}] (x) {$#1$};\draw(0,-1)--(x)--(0,1);}\,}
\newkeycommand{\dboxmap}[style=dbox][1]{\,\tikz{\node[style=\commandkey{style}] (x) {$#1$};\draw[boldedge] (0,-1.25)--(x)--(0,1.25);}\,}
\newkeycommand{\shortdboxmap}[style=dbox][1]{\,\tikz{\node[style=\commandkey{style}] (x) {$#1$};\draw[boldedge] (0,-1)--(x)--(0,1);}\,}

\newkeycommand{\maptypes}[style=map][3]{\begin{tikzpicture}
  \begin{pgfonlayer}{nodelayer}
    \node [style=\commandkey{style}] (0) at (0, 0) {$#1$};
    \node [style=none] (1) at (0, -1.5) {};
    \node [style=none] (2) at (0, 1.5) {};
    \node [style=right label] (3) at (0.25, -1.25) {$#2$};
    \node [style=right label] (4) at (0.25, 1) {$#3$};
  \end{pgfonlayer}
  \begin{pgfonlayer}{edgelayer}
    \draw (1.center) to (0);
    \draw (0) to (2.center);
  \end{pgfonlayer}
\end{tikzpicture}}

\newkeycommand{\wirelabel}[style=white label]{\,\tikz{\node[style=\commandkey{style}]{};}\,\xspace}

\newkeycommand{\pointmap}[style=point][1]{\,\tikz{\node[style=\commandkey{style}] (x) at (0,-0.3) {$#1$};\node [style=none] (3) at (0, 0.9) {};\node [style=none] (3) at (0, -0.9) {};\draw (x)--(0,0.7);}\,}

\newkeycommand{\point}[style=point,estyle=][1]{\,\tikz{\node[style=\commandkey{style}] (x) at (0,-0.05) {$#1$};\draw [\commandkey{estyle}] (x)--(0,1.05);}\,}

\newkeycommand{\copointmap}[style=copoint][1]{\,\tikz{\node[style=\commandkey{style}] (x) at (0,0.3) {$#1$};\draw (x)--(0,-0.7);}\,}

\newkeycommand{\dpoint}[style=point][1]{\,\tikz{\node[style=\commandkey{style},doubled] (x) at (0,-0.3) {$#1$};\draw[boldedge] (x)--(0,0.7);}\,}

\newkeycommand{\kpoint}[style=kpoint,estyle=][1]{\,\tikz{\node[style=\commandkey{style}] (x) at (0,-0.05) {$#1$};\draw [\commandkey{estyle}] (x)--(0,1.05);}\,}

\newkeycommand{\kpointgray}[style=gray kpoint,estyle=][1]{\,\tikz{\node[style=\commandkey{style}] (x) at (0,-0.05) {$#1$};\draw [\commandkey{estyle}] (x)--(0,1.05);}\,}

\newkeycommand{\typedkpoint}[style=kpoint,edgestyle=][2]{%
\,\begin{tikzpicture}
  \begin{pgfonlayer}{nodelayer}
    \node [style=none] (0) at (0, 1) {};
    \node [style=\commandkey{style}] (1) at (0, -0.75) {$#2$};
    \node [style=right label] (2) at (0.25, 0.5) {$#1$};
  \end{pgfonlayer}
  \begin{pgfonlayer}{edgelayer}
    \draw [\commandkey{edgestyle}] (1) to (0.center);
  \end{pgfonlayer}
\end{tikzpicture}\,}

\newkeycommand{\typedkpointdag}[style=kpoint adjoint,edgestyle=][2]{%
\,\begin{tikzpicture}
  \begin{pgfonlayer}{nodelayer}
    \node [style=none] (0) at (0, -1) {};
    \node [style=\commandkey{style}] (1) at (0, 0.75) {$#2$};
    \node [style=right label] (2) at (0.25, -0.5) {$#1$};
  \end{pgfonlayer}
  \begin{pgfonlayer}{edgelayer}
    \draw [\commandkey{edgestyle}] (0.center) to (1);
  \end{pgfonlayer}
\end{tikzpicture}\,}

\newkeycommand{\bistate}[style=kpoint][1]{\,\begin{tikzpicture}
  \begin{pgfonlayer}{nodelayer}
    \node [style=\commandkey{style}, minimum width=1 cm, inner sep=2pt] (0) at (0, -0.25) {$#1$};
    \node [style=none] (1) at (-0.75, 0) {};
    \node [style=none] (2) at (0.75, 0) {};
    \node [style=none] (3) at (-0.75, 0.88) {};
    \node [style=none] (4) at (0.75, 0.88) {};
  \end{pgfonlayer}
  \begin{pgfonlayer}{edgelayer}
    \draw (1.center) to (3.center);
    \draw (2.center) to (4.center);
  \end{pgfonlayer}
\end{tikzpicture}\,}

\newkeycommand{\bistatebig}[style=kpoint][1]{\,\begin{tikzpicture}
  \begin{pgfonlayer}{nodelayer}
    \node [style=\commandkey{style}, minimum width=, minimum width=1.26 cm, inner sep=2pt] (0) at (0, -0.25) {$#1$};
    \node [style=none] (1) at (-1, 0) {};
    \node [style=none] (2) at (1, 0) {};
    \node [style=none] (3) at (-1, 0.88) {};
    \node [style=none] (4) at (1, 0.88) {};
  \end{pgfonlayer}
  \begin{pgfonlayer}{edgelayer}
    \draw (1.center) to (3.center);
    \draw (2.center) to (4.center);
  \end{pgfonlayer}
\end{tikzpicture}\,}

\newkeycommand{\dbistate}[style=dkpoint][1]{\,\begin{tikzpicture}
  \begin{pgfonlayer}{nodelayer}
    \node [style=\commandkey{style}, minimum width=1 cm, inner sep=2pt] (0) at (0, -0.25) {$#1$};
    \node [style=none] (1) at (-0.75, 0) {};
    \node [style=none] (2) at (0.75, 0) {};
    \node [style=none] (3) at (-0.75, 1.0) {};
    \node [style=none] (4) at (0.75, 1.0) {};
  \end{pgfonlayer}
  \begin{pgfonlayer}{edgelayer}
    \draw [boldedge] (1.center) to (3.center);
    \draw [boldedge] (2.center) to (4.center);
  \end{pgfonlayer}
\end{tikzpicture}\,}

\newkeycommand{\bistatebraket}[style1=kpoint,style2=kpointadj][2]{\,\begin{tikzpicture}
  \begin{pgfonlayer}{nodelayer}
    \node [style=\commandkey{style1}, minimum width=1 cm, inner sep=2pt] (0) at (0, -0.75) {$#1$};
    \node [style=none] (1) at (-0.75, -0.5) {};
    \node [style=none] (2) at (0.75, -0.5) {};
    \node [style=none] (3) at (-0.75, 0.5) {};
    \node [style=none] (4) at (0.75, 0.5) {};
    \node [style=\commandkey{style2}, minimum width=1 cm, inner sep=2pt] (5) at (0, 0.75) {$#2$};
  \end{pgfonlayer}
  \begin{pgfonlayer}{edgelayer}
    \draw (1.center) to (3.center);
    \draw (2.center) to (4.center);
  \end{pgfonlayer}
\end{tikzpicture}\,}

\newkeycommand{\weight}[style=dkpoint][1]{\,\begin{tikzpicture}
  \begin{pgfonlayer}{nodelayer}
    \node [style=\commandkey{style}] (0) at (0, -0.5) {$#1$};
    \node [style=upground] (1) at (0, 0.75) {};
  \end{pgfonlayer}
  \begin{pgfonlayer}{edgelayer}
    \draw [style=boldedge] (0) to (1);
  \end{pgfonlayer}
\end{tikzpicture}\,}

\newkeycommand{\dkpoint}[style=dkpoint][1]{\,\tikz{\node[style=\commandkey{style}] (x) at (0,-0.1) {$#1$};\draw[boldedge] (x)--(0,1);}\,}
\newkeycommand{\dkpointadj}[style=dkpointadj][1]{\,\tikz{\node[style=\commandkey{style}] (x) at (0,0.1) {$#1$};\draw[boldedge] (x)--(0,-1);}\,}
\newkeycommand{\dkpointtrans}[style=dkpointtrans][1]{\,\tikz{\node[style=\commandkey{style}] (x) at (0,0.1) {$#1$};\draw[boldedge] (x)--(0,-1);}\,}
\newkeycommand{\dkpointconj}[style=dkpointconj][1]{\,\tikz{\node[style=\commandkey{style}] (x) at (0,-0.1) {$#1$};\draw[boldedge] (x)--(0,1);}\,}

\newkeycommand{\blackkpoint}[style=black kpoint][1]{\,\tikz{\node[style=\commandkey{style}] (x) at (0,-0.1) {$#1$};\draw (x)--(0,1);}\,}
\newkeycommand{\blackkpointadj}[style=black kpointadj][1]{\,\tikz{\node[style=\commandkey{style}] (x) at (0,0.1) {$#1$};\draw (x)--(0,-1);}\,}
\newkeycommand{\blackdkpoint}[style=black dkpoint][1]{\,\tikz{\node[style=\commandkey{style}] (x) at (0,-0.1) {$#1$};\draw[boldedge] (x)--(0,1);}\,}
\newkeycommand{\blackdkpointadj}[style=black dkpointadj][1]{\,\tikz{\node[style=\commandkey{style}] (x) at (0,0.1) {$#1$};\draw[boldedge] (x)--(0,-1);}\,}

\newcommand{\trace}{\,\begin{tikzpicture}
  \begin{pgfonlayer}{nodelayer}
    \node [style=none] (0) at (0, -0.60) {};
    \node [style=upground] (1) at (0, 0.40) {};
  \end{pgfonlayer}
  \begin{pgfonlayer}{edgelayer}
    \draw [style=boldedge] (0.center) to (1);
  \end{pgfonlayer}
\end{tikzpicture}\,}

\newcommand\discard\trace

\newkeycommand{\pointketbra}[style1=copoint,style2=point,estyle=][2]{\,%
\begin{tikzpicture}
  \begin{pgfonlayer}{nodelayer}
    \node [style=none] (0) at (0, -1.75) {};
    \node [style=\commandkey{style1}] (1) at (0, -0.75) {$#1$};
    \node [style=\commandkey{style2}] (2) at (0, 0.75) {$#2$};
    \node [style=none] (3) at (0, 1.75) {};
  \end{pgfonlayer}
  \begin{pgfonlayer}{edgelayer}
    \draw [\commandkey{estyle}] (0.center) to (1);
    \draw [\commandkey{estyle}] (2) to (3.center);
  \end{pgfonlayer}
\end{tikzpicture}\,}

\newkeycommand{\twocopointketbra}[style1=copoint,style2=copoint,style3=point][3]{\,%
\begin{tikzpicture}
  \begin{pgfonlayer}{nodelayer}
    \node [style=none] (0) at (-0.75, -1.75) {};
    \node [style=none] (1) at (0.75, -1.75) {};
    \node [style=\commandkey{style1}] (2) at (-0.75, -0.75) {$#1$};
    \node [style=\commandkey{style2}] (3) at (0.75, -0.75) {$#2$};
    \node [style=\commandkey{style3}] (4) at (0, 0.75) {$#3$};
    \node [style=none] (5) at (0, 1.75) {};
  \end{pgfonlayer}
  \begin{pgfonlayer}{edgelayer}
    \draw (0.center) to (2);
    \draw (1.center) to (3);
    \draw (4) to (5.center);
  \end{pgfonlayer}
\end{tikzpicture}\,}

\newkeycommand{\twopointketbra}[style1=copoint,style2=point,style3=point][3]{\,%
\begin{tikzpicture}
  \begin{pgfonlayer}{nodelayer}
    \node [style=none] (0) at (0, -1.75) {};
    \node [style=none] (1) at (-0.75, 1.75) {};
    \node [style=\commandkey{style1}] (2) at (0, -0.75) {$#1$};
    \node [style=\commandkey{style2}] (3) at (-0.75, 0.75) {$#2$};
    \node [style=\commandkey{style3}] (4) at (0.75, 0.75) {$#3$};
    \node [style=none] (5) at (0.75, 1.75) {};
  \end{pgfonlayer}
  \begin{pgfonlayer}{edgelayer}
    \draw (0.center) to (2);
    \draw (1.center) to (3);
    \draw (4) to (5.center);
  \end{pgfonlayer}
\end{tikzpicture}\,}

\newkeycommand{\pointbraket}[style1=point,style2=copoint,estyle=][2]{\,%
\begin{tikzpicture}
  \begin{pgfonlayer}{nodelayer}
    \node [style=\commandkey{style1}] (0) at (0, -0.625) {$#1$};
    \node [style=\commandkey{style2}] (1) at (0, 0.625) {$#2$};
  \end{pgfonlayer}
  \begin{pgfonlayer}{edgelayer}
    \draw [\commandkey{estyle}] (0) to (1);
  \end{pgfonlayer}
\end{tikzpicture}\,}

\newkeycommand{\kpointketbra}[style1=kpointdag,style2=kpoint,estyle=][2]{\,%
\begin{tikzpicture}
  \begin{pgfonlayer}{nodelayer}
    \node [style=none] (0) at (0, -1.9) {};
    \node [style=\commandkey{style1}] (1) at (0, -0.95) {$#1$};
    \node [style=\commandkey{style2}] (2) at (0, 0.95) {$#2$};
    \node [style=none] (3) at (0, 1.9) {};
  \end{pgfonlayer}
  \begin{pgfonlayer}{edgelayer}
    \draw [\commandkey{estyle}] (0.center) to (1);
    \draw [\commandkey{estyle}] (2) to (3.center);
  \end{pgfonlayer}
\end{tikzpicture}\,}

\newkeycommand{\kpointmap}[style1=kpoint,style2=map][2]{\,%
\begin{tikzpicture}
  \begin{pgfonlayer}{nodelayer}
    \node [style=\commandkey{style1}] (0) at (0, -1.1) {$#1$};
    \node [style=\commandkey{style2}] (1) at (0, 0.5) {$#2$};
    \node [style=none] (2) at (0, 1.75) {};
  \end{pgfonlayer}
  \begin{pgfonlayer}{edgelayer}
    \draw (0) to (1);
    \draw (1) to (2.center);
  \end{pgfonlayer}
\end{tikzpicture}%
\,}

\newkeycommand{\kpointmaptrans}[style1=kpointconj,style2=mapadj][2]{\,%
\begin{tikzpicture}
  \begin{pgfonlayer}{nodelayer}
    \node [style=\commandkey{style1}] (0) at (0, -1.1) {$#1$};
    \node [style=\commandkey{style2}] (1) at (0, 0.5) {$#2$};
    \node [style=none] (2) at (0, 1.75) {};
  \end{pgfonlayer}
  \begin{pgfonlayer}{edgelayer}
    \draw (0) to (1);
    \draw (1) to (2.center);
  \end{pgfonlayer}
\end{tikzpicture}%
\,}

\newkeycommand{\kpointmapadj}[style1=kpointadj,style2=map][2]{\,%
\begin{tikzpicture}
  \begin{pgfonlayer}{nodelayer}
    \node [style=\commandkey{style1}] (0) at (0, 1.1) {$#1$};
    \node [style=\commandkey{style2}] (1) at (0, -0.5) {$#2$};
    \node [style=none] (2) at (0, -1.75) {};
  \end{pgfonlayer}
  \begin{pgfonlayer}{edgelayer}
    \draw (0) to (1);
    \draw (1) to (2.center);
  \end{pgfonlayer}
\end{tikzpicture}%
\,}

\newkeycommand{\dkpointmap}[style1=dkpoint,style2=dmap][2]{\,%
\begin{tikzpicture}
  \begin{pgfonlayer}{nodelayer}
    \node [style=\commandkey{style1}] (0) at (0, -1.2) {$#1$};
    \node [style=\commandkey{style2}] (1) at (0, 0.4) {$#2$};
    \node [style=none] (2) at (0, 1.7) {};
  \end{pgfonlayer}
  \begin{pgfonlayer}{edgelayer}
    \draw[style=boldedge] (0) to (1);
    \draw[style=boldedge] (1) to (2.center);
  \end{pgfonlayer}
\end{tikzpicture}%
\,}

\newkeycommand{\dkpointmapx}[style1=dkpoint,style2=dmap][2]{\,%
\begin{tikzpicture}
  \begin{pgfonlayer}{nodelayer}
    \node [style=\commandkey{style1}] (0) at (0, -1.4) {$#1$};
    \node [style=\commandkey{style2}] (1) at (0, 0.6) {$#2$};
    \node [style=none] (2) at (0, 1.9) {};
  \end{pgfonlayer}
  \begin{pgfonlayer}{edgelayer}
    \draw[style=boldedge] (0) to (1);
    \draw[style=boldedge] (1) to (2.center);
  \end{pgfonlayer}
\end{tikzpicture}%
\,}

\newkeycommand{\boxcopointmapdag}[style1=map,style2=copoint][2]{\,%
\begin{tikzpicture}
  \begin{pgfonlayer}{nodelayer}
    \node [style=none] (0) at (0, -1.75) {};
    \node [style=\commandkey{style1}] (1) at (0, -0.5) {$#1$};
    \node [style=\commandkey{style2}] (2) at (0, 1) {$#2$};
  \end{pgfonlayer}
  \begin{pgfonlayer}{edgelayer}
    \draw (0.center) to (1);
    \draw (1) to (2);
  \end{pgfonlayer}
\end{tikzpicture}%
\,}

\newkeycommand{\kpointdagmap}[style1=map,style2=kpointdag][2]{\,%
\begin{tikzpicture}
  \begin{pgfonlayer}{nodelayer}
    \node [style=none] (0) at (0, -1.75) {};
    \node [style=\commandkey{style1}] (1) at (0, -0.5) {$#1$};
    \node [style=\commandkey{style2}] (2) at (0, 1.1) {$#2$};
  \end{pgfonlayer}
  \begin{pgfonlayer}{edgelayer}
    \draw (0.center) to (1);
    \draw (1) to (2);
  \end{pgfonlayer}
\end{tikzpicture}%
\,}

\newkeycommand{\sandwichmap}[style1=point,style2=map,style3=copoint][3]{\,%
\begin{tikzpicture}
  \begin{pgfonlayer}{nodelayer}
    \node [style=\commandkey{style1}] (0) at (0, -1.50) {$#1$};
    \node [style=\commandkey{style2}] (1) at (0, 0) {$#2$};
    \node [style=\commandkey{style3}] (2) at (0, 1.50) {$#3$};
  \end{pgfonlayer}
  \begin{pgfonlayer}{edgelayer}
    \draw (0) to (1);
    \draw (1) to (2);
  \end{pgfonlayer}
\end{tikzpicture}%
}

\newkeycommand{\kpointsandwichmap}[style1=kpoint,style2=map,style3=kpointdag][3]{\,%
\begin{tikzpicture}
  \begin{pgfonlayer}{nodelayer}
    \node [style=\commandkey{style1}] (0) at (0, -1.5) {$#1$};
    \node [style=\commandkey{style2}] (1) at (0, 0) {$#2$};
    \node [style=\commandkey{style3}] (2) at (0, 1.5) {$#3$};
  \end{pgfonlayer}
  \begin{pgfonlayer}{edgelayer}
    \draw (0) to (1);
    \draw (1) to (2);
  \end{pgfonlayer}
\end{tikzpicture}%
}

\newkeycommand{\sandwichmapconj}[style1=point,style2=mapconj,style3=copoint][3]{\,%
\begin{tikzpicture}
  \begin{pgfonlayer}{nodelayer}
    \node [style=\commandkey{style1}] (0) at (0, -1.50) {$#1$};
    \node [style=\commandkey{style2}] (1) at (0, 0) {$#2$};
    \node [style=\commandkey{style3}] (2) at (0, 1.50) {$#3$};
  \end{pgfonlayer}
  \begin{pgfonlayer}{edgelayer}
    \draw (0) to (1);
    \draw (1) to (2);
  \end{pgfonlayer}
\end{tikzpicture}%
}

\newkeycommand{\sandwichtwo}[style1=point,style2=map,style3=map,style4=copoint][4]{\,%
\begin{tikzpicture}
  \begin{pgfonlayer}{nodelayer}
    \node [style=\commandkey{style2}] (0) at (0, -1) {$#2$};
    \node [style=\commandkey{style3}] (1) at (0, 1) {$#3$};
    \node [style=\commandkey{style1}] (2) at (0, -2.6) {$#1$};
    \node [style=\commandkey{style4}] (3) at (0, 2.6) {$#4$};
  \end{pgfonlayer}
  \begin{pgfonlayer}{edgelayer}
    \draw (2) to (0);
    \draw (0) to (1);
    \draw (1) to (3);
  \end{pgfonlayer}
\end{tikzpicture}\,}

\newkeycommand{\longbraket}[style1=point,style2=copoint][2]{\,%
\begin{tikzpicture}
  \begin{pgfonlayer}{nodelayer}
    \node [style=\commandkey{style1}] (0) at (0, -2.6) {$#1$};
    \node [style=\commandkey{style2}] (1) at (0, 2.6) {$#2$};
  \end{pgfonlayer}
  \begin{pgfonlayer}{edgelayer}
    \draw (0) to (1);
  \end{pgfonlayer}
\end{tikzpicture}\,}

\newkeycommand{\onb}[style=point]{\ensuremath{\{\,\tikz{\node[style=\commandkey{style}] (x) at (0,-0.3) {$j$};\draw (x)--(0,0.7);}\,\}}\xspace}

\newkeycommand{\phasebraket}[style1=gray phase dot,style2=white phase dot][2]{\begin{tikzpicture}
  \begin{pgfonlayer}{nodelayer}
    \node [style=\commandkey{style1}] (0) at (0, -0.75) {$#1$};
    \node [style=\commandkey{style2}] (1) at (0, 0.5) {$#2$};
  \end{pgfonlayer}
  \begin{pgfonlayer}{edgelayer}
    \draw (1) to (0);
  \end{pgfonlayer}
\end{tikzpicture}}

\newkeycommand{\phase}[style=white phase dot][1]{\,\begin{tikzpicture}
    \begin{pgfonlayer}{nodelayer}
        \node [style=none] (0) at (0, 1) {};
        \node [style=\commandkey{style}] (2) at (0, -0) {$#1$}; 
        \node [style=none] (3) at (0, -1) {};
    \end{pgfonlayer}
    \begin{pgfonlayer}{edgelayer}
        \draw (2) to (0.center);
        \draw (3.center) to (2);
    \end{pgfonlayer}
\end{tikzpicture}\,}

\newkeycommand{\dphase}[style=white phase ddot][1]{\,\begin{tikzpicture}
    \begin{pgfonlayer}{nodelayer}
        \node [style=none] (0) at (0, 1) {};
        \node [style=\commandkey{style}] (2) at (0, -0) {$#1$};
        \node [style=none] (3) at (0, -1) {};
    \end{pgfonlayer}
    \begin{pgfonlayer}{edgelayer}
        \draw [boldedge] (2) to (0.center);
        \draw [boldedge] (3.center) to (2);
    \end{pgfonlayer}
\end{tikzpicture}\,}

\newkeycommand{\dphasepoint}[style=white phase ddot][1]{\,\begin{tikzpicture}[yshift=-3mm]
  \begin{pgfonlayer}{nodelayer}
    \node [style=none] (0) at (0, 1) {};
    \node [style=\commandkey{style}] (1) at (0, 0) {$#1$};
  \end{pgfonlayer}
  \begin{pgfonlayer}{edgelayer}
    \draw [style=boldedge] (0.center) to (1);
  \end{pgfonlayer}
\end{tikzpicture}\,}

\newkeycommand{\dphasecopoint}[style=white phase ddot][1]{\,\begin{tikzpicture}[yshift=3mm,yscale=-1]
  \begin{pgfonlayer}{nodelayer}
    \node [style=none] (0) at (0, 1) {};
    \node [style=\commandkey{style}] (1) at (0, 0) {$#1$};
  \end{pgfonlayer}
  \begin{pgfonlayer}{edgelayer}
    \draw [style=boldedge] (0.center) to (1);
  \end{pgfonlayer}
\end{tikzpicture}\,}

\newkeycommand{\dphasepointsm}[style=white phase ddot][1]{\,\begin{tikzpicture}[yshift=-3mm]
  \begin{pgfonlayer}{nodelayer}
    \node [style=none] (0) at (0, 1) {};
    \node [style=\commandkey{style}] (1) at (0, 0) {\footnotesize\!$#1$\!};
  \end{pgfonlayer}
  \begin{pgfonlayer}{edgelayer}
    \draw [style=boldedge] (0.center) to (1);
  \end{pgfonlayer}
\end{tikzpicture}\,}

\newkeycommand{\phasepoint}[style=white phase dot][1]{\,\begin{tikzpicture}
  \begin{pgfonlayer}{nodelayer}
    \node [style=none] (0) at (0, 0.75) {};
    \node [style=\commandkey{style}] (1) at (0, -0.25) {$#1$};
  \end{pgfonlayer}
  \begin{pgfonlayer}{edgelayer}
    \draw (0.center) to (1);
  \end{pgfonlayer}
\end{tikzpicture}\,}

\newkeycommand{\phasecopoint}[style=white phase dot][1]{\,\begin{tikzpicture}[yshift=3mm]
  \begin{pgfonlayer}{nodelayer}
    \node [style=none] (0) at (0, -0.75) {};
    \node [style=\commandkey{style}] (1) at (0, 0.25) {$#1$};
  \end{pgfonlayer}
  \begin{pgfonlayer}{edgelayer}
    \draw (0.center) to (1);
  \end{pgfonlayer}
\end{tikzpicture}\,}

\newkeycommand{\kpointbraket}[style1=kpoint,style2=kpoint adjoint,estyle=][2]{\,%
\begin{tikzpicture}
  \begin{pgfonlayer}{nodelayer}
    \node [style=\commandkey{style1}] (0) at (0, -0.735) {$#1$};
    \node [style=\commandkey{style2}] (1) at (0, 0.735) {$#2$};
  \end{pgfonlayer}
  \begin{pgfonlayer}{edgelayer}
    \draw [\commandkey{estyle}] (0) to (1);
  \end{pgfonlayer}
\end{tikzpicture}\,}

\newkeycommand{\kkpointbraket}[style1=kpoint,style2=kpoint adjoint,estyle=][2]{\,%
\begin{tikzpicture}
  \begin{pgfonlayer}{nodelayer}
    \node [style=\commandkey{style1}] (0) at (0, -0.685) {$#1$};
    \node [style=\commandkey{style2}] (1) at (0, 0.685) {$#2$};
  \end{pgfonlayer}
  \begin{pgfonlayer}{edgelayer}
    \draw [\commandkey{estyle}] (0) to (1);
  \end{pgfonlayer}
\end{tikzpicture}\,}

\newkeycommand{\kpointbraketx}[style1=kpoint,style2=kpoint adjoint,estyle=][2]{\,%
\begin{tikzpicture}
  \begin{pgfonlayer}{nodelayer}
    \node [style=\commandkey{style1}] (0) at (0, -0.885) {$#1$};
    \node [style=\commandkey{style2}] (1) at (0, 0.885) {$#2$};
  \end{pgfonlayer}
  \begin{pgfonlayer}{edgelayer}
    \draw [\commandkey{estyle}] (0) to (1);
  \end{pgfonlayer}
\end{tikzpicture}\,}

\tikzstyle{cdiag}=[matrix of math nodes, row sep=3em, column sep=3em, text height=1.5ex, text depth=0.25ex,inner sep=0.5em]
\tikzstyle{arrow above}=[transform canvas={yshift=0.5ex}]
\tikzstyle{arrow below}=[transform canvas={yshift=-0.5ex}]

\tikzset{
  col/.cd,
  0/.style={draw,fill=white}
  1/.style={draw,fill=blue}
}

\tikzstyle{B}=[draw,fill=gray!90!black]
\tikzstyle{W}=[draw,fill=white]
\tikzstyle{G}=[draw,fill=gray!50!white]

\tikzstyle{env}=[copoint,regular polygon rotate=0,minimum width=0.2cm, fill=black]

\tikzstyle{probs}=[shape=semicircle,fill=white,draw=black,shape border rotate=180,minimum width=1.2cm]

\tikzstyle{every picture}=[baseline=-0.25em,scale=0.5]  
\tikzstyle{dotpic}=[] 
\tikzstyle{diredges}=[every to/.style={diredge}]
\tikzstyle{math matrix}=[matrix of math nodes,left delimiter=(,right delimiter=),inner sep=2pt,column sep=1em,row sep=0.5em,nodes={inner sep=0pt},text height=1.5ex, text depth=0.25ex]

\tikzstyle{inline text}=[text height=1.5ex, text depth=0.25ex,yshift=0.5mm]
\tikzstyle{label}=[font=\footnotesize,text height=1.5ex, text depth=0.25ex,yshift=0.5mm]
\tikzstyle{left label}=[label,anchor=east,xshift=1.5mm]
\tikzstyle{right label}=[label,anchor=west,xshift=-1.5mm]

\tikzstyle{braceedge}=[decorate,decoration={brace,amplitude=2mm,raise=-1mm}]
\tikzstyle{small braceedge}=[decorate,decoration={brace,amplitude=1mm,raise=-1mm}]

\tikzstyle{doubled}=[line width=1.6pt] 
\tikzstyle{boldedge}=[doubled,shorten <=-0.17mm,shorten >=-0.17mm]
\tikzstyle{boldedgegray}=[doubled,gray,shorten <=-0.17mm,shorten >=-0.17mm]
\tikzstyle{singleedgegray}=[gray]

\tikzstyle{semidoubled}=[line width=1.4pt] 
\tikzstyle{semiboldedgegray}=[semidoubled,gray,shorten <=-0.17mm,shorten >=-0.17mm]

\tikzstyle{boxedge}=[semiboldedgegray]

\tikzstyle{boldedgedashed}=[very thick,dashed,shorten <=-0.17mm,shorten >=-0.17mm]
\tikzstyle{vboldedgedashed}=[doubled,dashed,shorten <=-0.17mm,shorten >=-0.17mm]
\tikzstyle{left hook arrow}=[left hook-latex]
\tikzstyle{right hook arrow}=[right hook-latex]
\tikzstyle{sembracket}=[line width=0.5pt,shorten <=-0.07mm,shorten >=-0.07mm]

\tikzstyle{causal edge}=[->,thick,gray]
\tikzstyle{causal nondir}=[thick,gray]
\tikzstyle{timeline}=[thick,gray, dashed]

\tikzstyle{cedge}=[<->,thick,gray!70!white]

\tikzstyle{empty diagram}=[draw=gray!40!white,dashed,shape=rectangle,minimum width=1cm,minimum height=1cm]
\tikzstyle{empty diagram small}=[draw=gray!50!white,dashed,shape=rectangle,minimum width=0.6cm,minimum height=0.5cm]

\tikzstyle{dot}=[inner sep=0mm,minimum width=2mm,minimum height=2mm,draw,shape=circle]  
\tikzstyle{Wsquare}=[white dot, shape=regular polygon, rounded corners=0.8 mm, minimum size=3.3 mm, regular polygon sides=3, outer sep=-0.2mm]
\tikzstyle{Wsquareadj}=[white dot, shape=regular polygon, rounded corners=0.8 mm, minimum size=3.3 mm, regular polygon sides=3, outer sep=-0.2mm, regular polygon rotate=180]
\tikzstyle{ddot}=[inner sep=0mm, doubled, minimum width=2.5mm,minimum height=2.5mm,draw,shape=circle]

\tikzstyle{black dot}=[dot,fill=black]
\tikzstyle{white dot}=[dot,fill=white,,text depth=-0.2mm]
\tikzstyle{white Wsquare}=[Wsquare,fill=white,,text depth=-0.2mm]
\tikzstyle{white Wsquareadj}=[Wsquareadj,fill=white,,text depth=-0.2mm]
\tikzstyle{green dot}=[white dot] 
\tikzstyle{gray dot}=[dot,fill=gray!40!white,,text depth=-0.2mm]
\tikzstyle{red dot}=[gray dot] 

\tikzstyle{black ddot}=[ddot,fill=black]
\tikzstyle{white ddot}=[ddot,fill=white]
\tikzstyle{gray ddot}=[ddot,fill=gray!40!white]

\tikzstyle{gray edge}=[gray!60!white]

\tikzstyle{small dot}=[inner sep=0.5mm,minimum width=0pt,minimum height=0pt,draw,shape=circle]

\tikzstyle{small black dot}=[small dot,fill=black]
\tikzstyle{small white dot}=[small dot,fill=white]
\tikzstyle{small gray dot}=[small dot,fill=gray!40!white]

\tikzstyle{causal dot}=[inner sep=0.4mm,minimum width=0pt,minimum height=0pt,draw=white,shape=circle,fill=gray!40!white]

\tikzstyle{phase dimensions}=[minimum size=5mm,font=\footnotesize,rectangle,rounded corners=2.5mm,inner sep=0.2mm,outer sep=-2mm]
\tikzstyle{dphase dimensions}=[minimum size=5mm,font=\footnotesize,rectangle,rounded corners=2.5mm,inner sep=0.2mm,outer sep=-2mm]

\tikzstyle{white phase dot}=[dot,fill=white,phase dimensions]
\tikzstyle{white phase ddot}=[ddot,fill=white,dphase dimensions]

\tikzstyle{white rect ddot}=[draw=black,fill=white,doubled,minimum size=5mm,font=\footnotesize,rectangle,rounded corners=2.5mm,inner sep=0.2mm]
\tikzstyle{gray rect ddot}=[draw=black,fill=gray!40!white,doubled,minimum size=6mm,font=\footnotesize,rectangle,rounded corners=3mm]

\tikzstyle{gray phase dot}=[dot,fill=gray!40!white,phase dimensions]
\tikzstyle{gray phase ddot}=[ddot,fill=gray!40!white,dphase dimensions]
\tikzstyle{grey phase dot}=[gray phase dot]
\tikzstyle{grey phase ddot}=[gray phase ddot]

\tikzstyle{small phase dimensions}=[minimum size=4mm,font=\tiny,rectangle,rounded corners=2mm,inner sep=0.2mm,outer sep=-2mm]
\tikzstyle{small dphase dimensions}=[minimum size=4mm,font=\tiny,rectangle,rounded corners=2mm,inner sep=0.2mm,outer sep=-2mm]

\tikzstyle{small gray phase dot}=[dot,fill=gray!40!white,small phase dimensions]
\tikzstyle{small gray phase ddot}=[ddot,fill=gray!40!white,small dphase dimensions]

\tikzstyle{small map}=[draw,shape=rectangle,minimum height=4mm,minimum width=4mm,fill=white]

\tikzstyle{cnot}=[fill=white,shape=circle,inner sep=-1.4pt]

\tikzstyle{asym hadamard}=[fill=white,draw,shape=NEbox,inner sep=0.6mm,font=\footnotesize,minimum height=4mm]
\tikzstyle{asym hadamard conj}=[fill=white,draw,shape=NWbox,inner sep=0.6mm,font=\footnotesize,minimum height=4mm]
\tikzstyle{asym hadamard dag}=[fill=white,draw,shape=SEbox,inner sep=0.6mm,font=\footnotesize,minimum height=4mm]

\tikzstyle{hadamard}=[fill=white,draw,inner sep=0.6mm,font=\footnotesize,minimum height=4mm,minimum width=4mm]
\tikzstyle{small hadamard}=[fill=white,draw,inner sep=0.6mm,minimum height=1.5mm,minimum width=1.5mm]
\tikzstyle{small hadamard rotate}=[small hadamard,rotate=45]
\tikzstyle{dhadamard}=[hadamard,doubled]
\tikzstyle{small dhadamard}=[small hadamard,doubled]
\tikzstyle{small dhadamard rotate}=[small hadamard rotate,doubled]
\tikzstyle{antipode}=[white dot,inner sep=0.3mm,font=\footnotesize]

\tikzstyle{scalar}=[diamond,draw,inner sep=0.5pt,font=\small]
\tikzstyle{dscalar}=[diamond,doubled, draw,inner sep=0.5pt,font=\small]

\tikzstyle{small box}=[rectangle,inline text,fill=white,draw,minimum height=5mm,yshift=-0.5mm,minimum width=5mm,font=\small]
\tikzstyle{small gray box}=[small box,fill=gray!30]
\tikzstyle{medium box}=[rectangle,inline text,fill=white,draw,minimum height=5mm,yshift=-0.5mm,minimum width=10mm,font=\small]
\tikzstyle{square box}=[small box] 
\tikzstyle{medium gray box}=[small box,fill=gray!30]
\tikzstyle{semilarge box}=[rectangle,inline text,fill=white,draw,minimum height=5mm,yshift=-0.5mm,minimum width=12.5mm,font=\small]
\tikzstyle{large box}=[rectangle,inline text,fill=white,draw,minimum height=5mm,yshift=-0.5mm,minimum width=15mm,font=\small]
\tikzstyle{large gray box}=[small box,fill=gray!30]

\tikzstyle{Bayes box}=[rectangle,fill=black,draw, minimum height=3mm, minimum width=3mm]

\tikzstyle{gray square point}=[small box,fill=gray!50]

\tikzstyle{dphase box white}=[dhadamard]
\tikzstyle{dphase box gray}=[dhadamard,fill=gray!50!white]
\tikzstyle{phase box white}=[hadamard]
\tikzstyle{phase box gray}=[hadamard,fill=gray!50!white]

\tikzstyle{point}=[regular polygon,regular polygon sides=3,draw,scale=0.75,inner sep=-0.5pt,minimum width=9mm,fill=white,regular polygon rotate=180]
\tikzstyle{point nosep}=[regular polygon,regular polygon sides=3,draw,scale=0.75,inner sep=-2pt,minimum width=9mm,fill=white,regular polygon rotate=180]
\tikzstyle{copoint}=[regular polygon,regular polygon sides=3,draw,scale=0.75,inner sep=-0.5pt,minimum width=9mm,fill=white]
\tikzstyle{dpoint}=[point,doubled]
\tikzstyle{dcopoint}=[copoint,doubled]

\tikzstyle{wide copoint}=[fill=white,draw,shape=isosceles triangle,shape border rotate=90,isosceles triangle stretches=true,inner sep=0pt,minimum width=1.5cm,minimum height=6.12mm]
\tikzstyle{wide point}=[fill=white,draw,shape=isosceles triangle,shape border rotate=-90,isosceles triangle stretches=true,inner sep=0pt,minimum width=1.5cm,minimum height=6.12mm,yshift=-0.0mm]
\tikzstyle{wide point plus}=[fill=white,draw,shape=isosceles triangle,shape border rotate=-90,isosceles triangle stretches=true,inner sep=0pt,minimum width=1.74cm,minimum height=7mm,yshift=-0.0mm]

\tikzstyle{wide dpoint}=[fill=white,doubled,draw,shape=isosceles triangle,shape border rotate=-90,isosceles triangle stretches=true,inner sep=0pt,minimum width=1.5cm,minimum height=6.12mm,yshift=-0.0mm]

\tikzstyle{tinypoint}=[regular polygon,regular polygon sides=3,draw,scale=0.55,inner sep=-0.15pt,minimum width=6mm,fill=white,regular polygon rotate=180] 

\tikzstyle{white point}=[point]
\tikzstyle{white dpoint}=[dpoint]
\tikzstyle{green point}=[white point] 
\tikzstyle{white copoint}=[copoint]
\tikzstyle{gray point}=[point,fill=gray!40!white]
\tikzstyle{gray dpoint}=[gray point,doubled]
\tikzstyle{red point}=[gray point] 
\tikzstyle{gray copoint}=[copoint,fill=gray!40!white]
\tikzstyle{gray dcopoint}=[gray copoint,doubled]

\tikzstyle{white point guide}=[regular polygon,regular polygon sides=3,font=\scriptsize,draw,scale=0.65,inner sep=-0.5pt,minimum width=9mm,fill=white,regular polygon rotate=180]

\tikzstyle{black point}=[point,fill=black,font=\color{white}]
\tikzstyle{black copoint}=[copoint,fill=black,font=\color{white}]

\tikzstyle{tiny gray point}=[tinypoint,fill=gray!40!white]

\tikzstyle{diredge}=[->]
\tikzstyle{ddiredge}=[<->]
\tikzstyle{rdiredge}=[<-]
\tikzstyle{thickdiredge}=[->, very thick]
\tikzstyle{pointer edge}=[->,very thick,gray]
\tikzstyle{pointer edge part}=[very thick,gray]
\tikzstyle{dashed edge}=[dashed]
\tikzstyle{thick dashed edge}=[very thick,dashed]
\tikzstyle{thick gray dashed edge}=[thick dashed edge,gray!40]
\tikzstyle{thick map edge}=[very thick,|->]

\makeatletter
\newcommand{\boxshape}[3]{%
\pgfdeclareshape{#1}{
\inheritsavedanchors[from=rectangle] 
\inheritanchorborder[from=rectangle]
\inheritanchor[from=rectangle]{center}
\inheritanchor[from=rectangle]{north}
\inheritanchor[from=rectangle]{south}
\inheritanchor[from=rectangle]{west}
\inheritanchor[from=rectangle]{east}
\backgroundpath{
\southwest \pgf@xa=\pgf@x \pgf@ya=\pgf@y
\northeast \pgf@xb=\pgf@x \pgf@yb=\pgf@y

\@tempdima=#2
\@tempdimb=#3

\pgfpathmoveto{\pgfpoint{\pgf@xa - 5pt + \@tempdima}{\pgf@ya}}
\pgfpathlineto{\pgfpoint{\pgf@xa - 5pt - \@tempdima}{\pgf@yb}}
\pgfpathlineto{\pgfpoint{\pgf@xb + 5pt + \@tempdimb}{\pgf@yb}}
\pgfpathlineto{\pgfpoint{\pgf@xb + 5pt - \@tempdimb}{\pgf@ya}}
\pgfpathlineto{\pgfpoint{\pgf@xa - 5pt + \@tempdima}{\pgf@ya}}
\pgfpathclose
}
}}

\boxshape{NEbox}{0pt}{5pt}
\boxshape{SEbox}{0pt}{-5pt}
\boxshape{NWbox}{5pt}{0pt}
\boxshape{SWbox}{-5pt}{0pt}
\boxshape{EBox}{-3pt}{3pt}
\boxshape{WBox}{3pt}{-3pt}
\makeatother

\tikzstyle{cloud}=[shape=cloud,draw,minimum width=1.5cm,minimum height=1.5cm]

\tikzstyle{map}=[draw,shape=NEbox,inner sep=2pt,minimum height=6mm,fill=white]
\tikzstyle{dashedmap}=[draw,dashed,shape=NEbox,inner sep=2pt,minimum height=6mm,fill=white]
\tikzstyle{mapdag}=[draw,shape=SEbox,inner sep=2pt,minimum height=6mm,fill=white]
\tikzstyle{mapadj}=[draw,shape=SEbox,inner sep=2pt,minimum height=6mm,fill=white]
\tikzstyle{maptrans}=[draw,shape=SWbox,inner sep=2pt,minimum height=6mm,fill=white]
\tikzstyle{mapconj}=[draw,shape=NWbox,inner sep=2pt,minimum height=6mm,fill=white]

\tikzstyle{medium map}=[draw,shape=NEbox,inner sep=2pt,minimum height=6mm,fill=white,minimum width=7mm]
\tikzstyle{medium map dag}=[draw,shape=SEbox,inner sep=2pt,minimum height=6mm,fill=white,minimum width=7mm]
\tikzstyle{medium map adj}=[draw,shape=SEbox,inner sep=2pt,minimum height=6mm,fill=white,minimum width=7mm]
\tikzstyle{medium map trans}=[draw,shape=SWbox,inner sep=2pt,minimum height=6mm,fill=white,minimum width=7mm]
\tikzstyle{medium map conj}=[draw,shape=NWbox,inner sep=2pt,minimum height=6mm,fill=white,minimum width=7mm]
\tikzstyle{semilarge map}=[draw,shape=NEbox,inner sep=2pt,minimum height=6mm,fill=white,minimum width=9.5mm]
\tikzstyle{semilarge map trans}=[draw,shape=SWbox,inner sep=2pt,minimum height=6mm,fill=white,minimum width=9.5mm]
\tikzstyle{semilarge map adj}=[draw,shape=SEbox,inner sep=2pt,minimum height=6mm,fill=white,minimum width=9.5mm]
\tikzstyle{semilarge map dag}=[draw,shape=SEbox,inner sep=2pt,minimum height=6mm,fill=white,minimum width=9.5mm]
\tikzstyle{semilarge map conj}=[draw,shape=NWbox,inner sep=2pt,minimum height=6mm,fill=white,minimum width=9.5mm]
\tikzstyle{large map}=[draw,shape=NEbox,inner sep=2pt,minimum height=6mm,fill=white,minimum width=12mm]
\tikzstyle{large map conj}=[draw,shape=NWbox,inner sep=2pt,minimum height=6mm,fill=white,minimum width=12mm]
\tikzstyle{very large map}=[draw,shape=NEbox,inner sep=2pt,minimum height=6mm,fill=white,minimum width=17mm]

\tikzstyle{medium dmap}=[draw,doubled,shape=NEbox,inner sep=2pt,minimum height=6mm,fill=white,minimum width=7mm]
\tikzstyle{medium dmap dag}=[draw,doubled,shape=SEbox,inner sep=2pt,minimum height=6mm,fill=white,minimum width=7mm]
\tikzstyle{medium dmap adj}=[draw,doubled,shape=SEbox,inner sep=2pt,minimum height=6mm,fill=white,minimum width=7mm]
\tikzstyle{medium dmap trans}=[draw,doubled,shape=SWbox,inner sep=2pt,minimum height=6mm,fill=white,minimum width=7mm]
\tikzstyle{medium dmap conj}=[draw,doubled,shape=NWbox,inner sep=2pt,minimum height=6mm,fill=white,minimum width=7mm]
\tikzstyle{semilarge dmap}=[draw,doubled,shape=NEbox,inner sep=2pt,minimum height=6mm,fill=white,minimum width=9.5mm]
\tikzstyle{semilarge dmap trans}=[draw,doubled,shape=SWbox,inner sep=2pt,minimum height=6mm,fill=white,minimum width=9.5mm]
\tikzstyle{semilarge dmap adj}=[draw,doubled,shape=SEbox,inner sep=2pt,minimum height=6mm,fill=white,minimum width=9.5mm]
\tikzstyle{semilarge dmap dag}=[draw,doubled,shape=SEbox,inner sep=2pt,minimum height=6mm,fill=white,minimum width=9.5mm]
\tikzstyle{semilarge dmap conj}=[draw,doubled,shape=NWbox,inner sep=2pt,minimum height=6mm,fill=white,minimum width=9.5mm]
\tikzstyle{large dmap}=[draw,doubled,shape=NEbox,inner sep=2pt,minimum height=6mm,fill=white,minimum width=12mm]
\tikzstyle{large dmap conj}=[draw,doubled,shape=NWbox,inner sep=2pt,minimum height=6mm,fill=white,minimum width=12mm]
\tikzstyle{large dmap trans}=[draw,doubled,shape=SWbox,inner sep=2pt,minimum height=6mm,fill=white,minimum width=12mm]
\tikzstyle{large dmap adj}=[draw,doubled,shape=SEbox,inner sep=2pt,minimum height=6mm,fill=white,minimum width=12mm]
\tikzstyle{large dmap dag}=[draw,doubled,shape=SEbox,inner sep=2pt,minimum height=6mm,fill=white,minimum width=12mm]
\tikzstyle{very large dmap}=[draw,doubled,shape=NEbox,inner sep=2pt,minimum height=6mm,fill=white,minimum width=19.5mm]

\tikzstyle{muxbox}=[draw,shape=rectangle,minimum height=3mm,minimum width=3mm,fill=white]
\tikzstyle{dmuxbox}=[muxbox,doubled]

\tikzstyle{box}=[draw,shape=rectangle,inner sep=2pt,minimum height=6mm,minimum width=6mm,fill=white]
\tikzstyle{dbox}=[draw,doubled,shape=rectangle,inner sep=2pt,minimum height=6mm,minimum width=6mm,fill=white]
\tikzstyle{dmap}=[draw,doubled,shape=NEbox,inner sep=2pt,minimum height=6mm,fill=white]
\tikzstyle{dmapdag}=[draw,doubled,shape=SEbox,inner sep=2pt,minimum height=6mm,fill=white]
\tikzstyle{dmapadj}=[draw,doubled,shape=SEbox,inner sep=2pt,minimum height=6mm,fill=white]
\tikzstyle{dmaptrans}=[draw,doubled,shape=SWbox,inner sep=2pt,minimum height=6mm,fill=white]
\tikzstyle{dmapconj}=[draw,doubled,shape=NWbox,inner sep=2pt,minimum height=6mm,fill=white]

\tikzstyle{ddmap}=[draw,doubled,dashed,shape=NEbox,inner sep=2pt,minimum height=6mm,fill=white]
\tikzstyle{ddmapdag}=[draw,doubled,dashed,shape=SEbox,inner sep=2pt,minimum height=6mm,fill=white]
\tikzstyle{ddmapadj}=[draw,doubled,dashed,shape=SEbox,inner sep=2pt,minimum height=6mm,fill=white]
\tikzstyle{ddmaptrans}=[draw,doubled,dashed,shape=SWbox,inner sep=2pt,minimum height=6mm,fill=white]
\tikzstyle{ddmapconj}=[draw,doubled,dashed,shape=NWbox,inner sep=2pt,minimum height=6mm,fill=white]

\boxshape{sNEbox}{0pt}{3pt}
\boxshape{sSEbox}{0pt}{-3pt}
\boxshape{sNWbox}{3pt}{0pt}
\boxshape{sSWbox}{-3pt}{0pt}
\tikzstyle{smap}=[draw,shape=sNEbox,fill=white]
\tikzstyle{smapdag}=[draw,shape=sSEbox,fill=white]
\tikzstyle{smapadj}=[draw,shape=sSEbox,fill=white]
\tikzstyle{smaptrans}=[draw,shape=sSWbox,fill=white]
\tikzstyle{smapconj}=[draw,shape=sNWbox,fill=white]

\tikzstyle{dsmap}=[draw,dashed,shape=sNEbox,fill=white]
\tikzstyle{dsmapdag}=[draw,dashed,shape=sSEbox,fill=white]
\tikzstyle{dsmaptrans}=[draw,dashed,shape=sSWbox,fill=white]
\tikzstyle{dsmapconj}=[draw,dashed,shape=sNWbox,fill=white]

\boxshape{mNEbox}{0pt}{10pt}
\boxshape{mSEbox}{0pt}{-10pt}
\boxshape{mNWbox}{10pt}{0pt}
\boxshape{mSWbox}{-10pt}{0pt}
\tikzstyle{mmap}=[draw,shape=mNEbox]
\tikzstyle{mmapdag}=[draw,shape=mSEbox]
\tikzstyle{mmaptrans}=[draw,shape=mSWbox]
\tikzstyle{mmapconj}=[draw,shape=mNWbox]

\tikzstyle{mmapgray}=[draw,fill=gray!40!white,shape=mNEbox]
\tikzstyle{smapgray}=[draw,fill=gray!40!white,shape=sNEbox]

\makeatletter

\pgfdeclareshape{cornerpoint}{
\inheritsavedanchors[from=rectangle] 
\inheritanchorborder[from=rectangle]
\inheritanchor[from=rectangle]{center}
\inheritanchor[from=rectangle]{north}
\inheritanchor[from=rectangle]{south}
\inheritanchor[from=rectangle]{west}
\inheritanchor[from=rectangle]{east}
\backgroundpath{
\southwest \pgf@xa=\pgf@x \pgf@ya=\pgf@y
\northeast \pgf@xb=\pgf@x \pgf@yb=\pgf@y

\pgfmathsetmacro{\pgf@shorten@left}{\pgfkeysvalueof{/tikz/shorten left}}
\pgfmathsetmacro{\pgf@shorten@right}{\pgfkeysvalueof{/tikz/shorten right}}

\pgfpathmoveto{\pgfpoint{0.5 * (\pgf@xa + \pgf@xb)}{\pgf@ya - 5pt}}
\pgfpathlineto{\pgfpoint{\pgf@xa - 8pt + \pgf@shorten@left}{\pgf@yb - 1.5 * \pgf@shorten@left}}
\pgfpathlineto{\pgfpoint{\pgf@xa - 8pt + \pgf@shorten@left}{\pgf@yb}}
\pgfpathlineto{\pgfpoint{\pgf@xb + 8pt - \pgf@shorten@right}{\pgf@yb}}
\pgfpathlineto{\pgfpoint{\pgf@xb + 8pt - \pgf@shorten@right}{\pgf@yb - 1.5 * \pgf@shorten@right}}
\pgfpathclose
}
}

\pgfdeclareshape{cornercopoint}{
\inheritsavedanchors[from=rectangle] 
\inheritanchorborder[from=rectangle]
\inheritanchor[from=rectangle]{center}
\inheritanchor[from=rectangle]{north}
\inheritanchor[from=rectangle]{south}
\inheritanchor[from=rectangle]{west}
\inheritanchor[from=rectangle]{east}
\backgroundpath{
\southwest \pgf@xa=\pgf@x \pgf@ya=\pgf@y
\northeast \pgf@xb=\pgf@x \pgf@yb=\pgf@y

\pgfmathsetmacro{\pgf@shorten@left}{\pgfkeysvalueof{/tikz/shorten left}}
\pgfmathsetmacro{\pgf@shorten@right}{\pgfkeysvalueof{/tikz/shorten right}}

\pgfpathmoveto{\pgfpoint{0.5 * (\pgf@xa + \pgf@xb)}{\pgf@yb + 5pt}}
\pgfpathlineto{\pgfpoint{\pgf@xa - 8pt + \pgf@shorten@left}{\pgf@ya + 1.5 * \pgf@shorten@left}}
\pgfpathlineto{\pgfpoint{\pgf@xa - 8pt + \pgf@shorten@left}{\pgf@ya}}
\pgfpathlineto{\pgfpoint{\pgf@xb + 8pt - \pgf@shorten@right}{\pgf@ya}}
\pgfpathlineto{\pgfpoint{\pgf@xb + 8pt - \pgf@shorten@right}{\pgf@ya + 1.5 * \pgf@shorten@right}}
\pgfpathclose
}
}

\makeatother

\pgfkeyssetvalue{/tikz/shorten left}{0pt}
\pgfkeyssetvalue{/tikz/shorten right}{0pt}

\tikzstyle{kpoint common}=[draw,fill=white,inner sep=1pt,minimum height=4mm]
\tikzstyle{kpoint sc}=[shape=cornerpoint,kpoint common]
\tikzstyle{kpoint adjoint sc}=[shape=cornercopoint,kpoint common]
\tikzstyle{kpoint}=[shape=cornerpoint,shorten left=5pt,kpoint common]
\tikzstyle{kpoint adjoint}=[shape=cornercopoint,shorten left=5pt,kpoint common]
\tikzstyle{kpoint conjugate}=[shape=cornerpoint,shorten right=5pt,kpoint common]
\tikzstyle{kpoint transpose}=[shape=cornercopoint,shorten right=5pt,kpoint common]
\tikzstyle{kpoint symm}=[shape=cornerpoint,shorten left=5pt,shorten right=5pt,kpoint common]

\tikzstyle{wide kpoint sc}=[shape=cornerpoint,kpoint common, minimum width=1 cm]
\tikzstyle{wide kpointdag sc}=[shape=cornercopoint,kpoint common, minimum width=1 cm]

\tikzstyle{black kpoint}=[shape=cornerpoint,shorten left=5pt,kpoint common,fill=black,font=\color{white}]
\tikzstyle{black kpoint adjoint}=[shape=cornercopoint,shorten left=5pt,kpoint common,fill=black,font=\color{white}]
\tikzstyle{black kpointadj}=[shape=cornercopoint,shorten left=5pt,kpoint common,fill=black,font=\color{white}]

\tikzstyle{black dkpoint}=[shape=cornerpoint,shorten left=5pt,kpoint common,fill=black, doubled,font=\color{white}]
\tikzstyle{black dkpoint adjoint}=[shape=cornercopoint,shorten left=5pt,kpoint common,fill=black, doubled,font=\color{white}]
\tikzstyle{black dkpointadj}=[shape=cornercopoint,shorten left=5pt,kpoint common,fill=black, doubled,font=\color{white}] 

\tikzstyle{kpointdag}=[kpoint adjoint]
\tikzstyle{kpointadj}=[kpoint adjoint]
\tikzstyle{kpointconj}=[kpoint conjugate]
\tikzstyle{kpointtrans}=[kpoint transpose]

\tikzstyle{big kpoint}=[kpoint, minimum width=1.2 cm, minimum height=8mm, inner sep=4pt, text depth=3mm]

\tikzstyle{wide kpoint}=[kpoint, minimum width=1 cm, inner sep=2pt]
\tikzstyle{wide kpointdag}=[kpointdag, minimum width=1 cm, inner sep=2pt]
\tikzstyle{wide kpointconj}=[kpointconj, minimum width=1 cm, inner sep=2pt]
\tikzstyle{wide kpointtrans}=[kpointtrans, minimum width=1 cm, inner sep=2pt]

\tikzstyle{wider kpoint}=[kpoint, minimum width=1.25 cm, inner sep=2pt]
\tikzstyle{wider kpointdag}=[kpointdag, minimum width=1.25 cm, inner sep=2pt]
\tikzstyle{wider kpointconj}=[kpointconj, minimum width=1.25 cm, inner sep=2pt]
\tikzstyle{wider kpointtrans}=[kpointtrans, minimum width=1.25 cm, inner sep=2pt]

\tikzstyle{gray kpoint}=[kpoint,fill=gray!50!white]
\tikzstyle{gray kpointdag}=[kpointdag,fill=gray!50!white]
\tikzstyle{gray kpointadj}=[kpointadj,fill=gray!50!white]
\tikzstyle{gray kpointconj}=[kpointconj,fill=gray!50!white]
\tikzstyle{gray kpointtrans}=[kpointtrans,fill=gray!50!white]

\tikzstyle{gray dkpoint}=[kpoint,fill=gray!50!white,doubled]
\tikzstyle{gray dkpointdag}=[kpointdag,fill=gray!50!white,doubled]
\tikzstyle{gray dkpointadj}=[kpointadj,fill=gray!50!white,doubled]
\tikzstyle{gray dkpointconj}=[kpointconj,fill=gray!50!white,doubled]
\tikzstyle{gray dkpointtrans}=[kpointtrans,fill=gray!50!white,doubled]

\tikzstyle{white label}=[draw,fill=white,rectangle,inner sep=0.7 mm]
\tikzstyle{gray label}=[draw,fill=gray!50!white,rectangle,inner sep=0.7 mm]
\tikzstyle{black label}=[draw,fill=black,rectangle,inner sep=0.7 mm]

\tikzstyle{dkpoint}=[kpoint,doubled]
\tikzstyle{wide dkpoint}=[wide kpoint,doubled]
\tikzstyle{dkpointdag}=[kpoint adjoint,doubled]
\tikzstyle{wide dkpointdag}=[wide kpointdag,doubled]
\tikzstyle{dkcopoint}=[kpoint adjoint,doubled]
\tikzstyle{dkpointadj}=[kpoint adjoint,doubled]
\tikzstyle{dkpointconj}=[kpoint conjugate,doubled]
\tikzstyle{dkpointtrans}=[kpoint transpose,doubled]

\tikzstyle{kscalar}=[kpoint common, shape=EBox, inner xsep=-1pt, inner ysep=3pt,font=\small]
\tikzstyle{kscalarconj}=[kpoint common, shape=WBox, inner xsep=-1pt, inner ysep=3pt,font=\small]

\tikzstyle{spekpoint}=[kpoint sc,minimum height=5mm,inner sep=3pt]
\tikzstyle{spekcopoint}=[kpoint adjoint sc,minimum height=5mm,inner sep=3pt]

\tikzstyle{dspekpoint}=[spekpoint,doubled]
\tikzstyle{dspekcopoint}=[spekcopoint,doubled]

 \tikzstyle{upground}=[circuit ee IEC,thick,ground,rotate=90,scale=2.5]
 \tikzstyle{downground}=[circuit ee IEC,thick,ground,rotate=-90,scale=2.5]
 \tikzstyle{bigground}=[regular polygon,regular polygon sides=3,draw=gray,scale=0.50,inner sep=-0.5pt,minimum width=10mm,fill=gray]

\tikzstyle{arrs}=[-latex,font=\small,auto]
\tikzstyle{arrow plain}=[arrs]
\tikzstyle{arrow dashed}=[dashed,arrs]
\tikzstyle{arrow bold}=[very thick,arrs]
\tikzstyle{arrow hide}=[draw=white!0,-]
\tikzstyle{arrow reverse}=[latex-]
\tikzstyle{cdnode}=[]

\let\olddagger\dagger
\renewcommand{\dagger}{\ensuremath{\olddagger}\xspace}

\usepackage{makeidx}
\makeindex

\theoremstyle{definition}
\newtheorem{theorem}{Theorem}[section]
\newtheorem*{theorem*}{Theorem}

\newtheorem{example*}[theorem]{Example*}
\newtheorem{examples*}[theorem]{Examples*}
\newtheorem{remark}[theorem]{Remark}
\newtheorem{remark*}[theorem]{Remark*}

\if\lecturenotes0

\newtheorem{exer}[theorem]{Exercise}
\newtheorem{exeropt}[theorem]{Exercise}
\newtheorem{exer*}[theorem]{Exercise*}

\fi

\if\lecturenotes1

\newtheorem{exer*}[theorem]{Exercise*}

\newtheoremstyle{exercise}{3pt}{3pt}{\color{red}}{}{\bf}{}{.5em}{}
\theoremstyle{exercise}

\fi

\newcommand{\TODO}[1]{\marginpar{\scriptsize\bB \textbf{TODO:} #1\e}}

\newcommand{\TODOa}[1]{\marginpar{\scriptsize\bM \textbf{TODO:} #1\e}}
\newcommand{\TODOb}[1]{\marginpar{\scriptsize\bB \textbf{TODO:} #1\e}}

\newcommand{\COMMa}[1]{\marginpar{\scriptsize\bM \textbf{COMM:} #1\e}}
\newcommand{\COMMb}[1]{\marginpar{\scriptsize\bB \textbf{COMM:} #1\e}}

\newcommand{\CHECK}[1]{\marginpar{\scriptsize\bR \textbf{CHECK:} #1\e}}

\newkeycommand{\moral}[width=11cm][1]{\begin{center}
\fbox{\ \parbox{\commandkey{width}}{\centering #1\vphantom{Xy}}\ } 
\end{center}}

\newkeycommand{\morallong}[width=11cm][1]{\par\medskip\noindent
\centerline{\fbox{\ \parbox{\commandkey{width}}{\centering #1\vphantom{Xy}}\ }} 
\par\medskip\noindent}

\usepackage{color}
\def\bR{\begin{color}{red}} 
\def\bB{\begin{color}{blue}}
\def\bM{\begin{color}{magenta}}
\def\bC{\begin{color}{cyan}}
\def\bW{\begin{color}{white}}
\def\bBl{\begin{color}{black}} 
\def\bG{\begin{color}{green}}
\def\bY{\begin{color}{yellow}}
\def\e{\end{color}\xspace}
\newcommand{\bit}{\begin{itemize}}
\newcommand{\eit}{\end{itemize}\par\noindent}
\newcommand{\ben}{\begin{enumerate}}
\newcommand{\een}{\end{enumerate}\par\noindent}
\newcommand{\beq}{\begin{equation}}
\newcommand{\eeq}{\end{equation}\par\noindent}
\newcommand{\beqa}{\begin{eqnarray*}}
\newcommand{\eeqa}{\end{eqnarray*}\par\noindent}
\newcommand{\beqn}{\begin{eqnarray}}
\newcommand{\eeqn}{\end{eqnarray}\par\noindent}

\if\lecturenotes1

\renewcommand{\TODO}[1]{}
\renewcommand{\TODOa}[1]{}
\renewcommand{\TODOb}[1]{}
\renewcommand{\COMMa}[1]{}
\renewcommand{\COMMb}[1]{}
\renewcommand{\CHECK}[1]{}

\def\bR{\begin{color}{black}} 
\def\bB{\begin{color}{black}}
\def\bM{\begin{color}{black}}
\def\bC{\begin{color}{black}}
\def\bW{\begin{color}{black}}
\def\bG{\begin{color}{black}}
\def\bY{\begin{color}{black}}

\fi

\if\cupversion1

\renewcommand{\TODO}[1]{}
\renewcommand{\TODOa}[1]{}
\renewcommand{\TODOb}[1]{}
\renewcommand{\COMMa}[1]{}
\renewcommand{\COMMb}[1]{}
\renewcommand{\CHECK}[1]{}

\def\bR{\begin{color}{black}} 
\def\bB{\begin{color}{black}}
\def\bM{\begin{color}{black}}
\def\bC{\begin{color}{black}}
\def\bW{\begin{color}{black}}
\def\bG{\begin{color}{black}}
\def\bY{\begin{color}{black}}

\fi

\usepackage{color}
\def\bR{\begin{color}{red}}     
\def\bB{\begin{color}{blue}}
\def\bM{\begin{color}{magenta}}  
\def\bC{\begin{color}{cyan}}  
\def\bW{\begin{color}{white}}
\def\bBl{\begin{color}{black}}   
\def\bG{\begin{color}{green}}
\def\bY{\begin{color}{yellow}}  
\def\e{\end{color}\xspace}

\providecommand{\urlalt}[2]{\href{#1}{#2}}

\title{Dual Density Operators and Natural Language Meaning}  

\author{Daniela Ashoush and Bob Coecke
  \institute{University of Oxford\\Department of Computer Science}
  \email{dan.ashoush@gmail.com - bob.coecke@cs.ox.ac.uk}      
  }

\def\titlerunning{Dual Density Operators and Natural Language Meaning}       
\def\authorrunning{D.~Ashoush \& B.~Coecke}   

\begin{document}
  
\maketitle

\begin{abstract}
Density operators allow for representing ambiguity about a vector representation, both in quantum theory and in  distributional natural language meaning. Formally equivalently, they allow for discarding part of the description of a composite system, where we consider the discarded part to be the context. We introduce dual density operators, which allow for two independent notions of context. We demonstrate the use of dual density operators within a grammatical-compositional distributional framework for natural language meaning. We show that dual density operators can be used to simultaneously represent: (i) ambiguity about word meanings (e.g.~queen as a person vs.~queen as a band), and (ii) lexical entailment (e.g.~tiger  $\Rightarrow$ mammal).  We provide a proof-of-concept example.  
\end{abstract}

\section{Introduction}    
 
In \cite{vN} von Neumann introduced \em density operators \em in order to give a description of quantum systems for which we don't have perfect knowledge about their state, but rather, there is a probability distribution describing the likeliness to be in a each state.  The result is not a standard probability distribution, but one that also accounts for the `probabilistic distance' between vectors as given by the Born-rule, i.e.~the square-modulus of the inner-product  \cite{Gleason}.    

However, vectors are not only used to represent the states of quantum systems.  In natural language processing (NLP) they are also used to represent the meanings of words \cite{lundburgess1996, Schuetze}, and (some function of) the inner-product is typically taken to be a similarity measure.  However, the meanings of many words are ambiguous, that is,  the same word is used to describe very different things: ``queen'' can be a monarch, a rock band, a bee, or a chess piece.  Also in this case it is very natural to use density operators  in order to allow for a lack of knowledge on which meaning (i.e.~which vector) is intended \cite{DimitriDPhil, RobinMSc, calco2015}.   Since density operators admit `mixing', we can now mix the vectors representing the distinct meanings of an ambiguous word into a density operator representing that ambiguous word:  
\begin{center}
queen := ${1\over 4}$ (queen-monarch + queen-bee + queen-band + queen-chess)      
\end{center}

Besides accounting for similarity of words, a vector representation of word meanings also allows for compositional reasoning: given the grammatical structure of a phrase or a sentence and the meanings of the words therein, one can compute the meaning  of that phrase or sentence \cite{CSC, GrefSadr, KartSadr}.  The crux to doing so is the fact that vectors  inhabit a category of a structure that matches the structure  of grammar  \cite{LambekvsLambek}, resulting in meanings of words being `teleported' through a sentence \cite{teleling}.  

Moreover, this algorithm for computing phrase and sentence meanings from word meanings carries over to density operators, since via Selinger's CPM-construction \cite{SelingerCPM}  these also inhabit  a category of the appropriate structure.   It is indeed an important feature of the framework of  \cite{CSC} that it is not attached to a particular representation of word-meanings. The passage to density operators also allows for retaining standard empirical methods, hence resulting in data-driven and grammar-driven compositional reasoning about ambiguous words \cite{calco2015}.  This allows one, for example, to observe how the ambiguity (measured by e.g.~von Neumann entropy) vanishes thanks to the disambiguating role other words play in the sentence.  

Now, ambiguity is not the only feature of natural language that is not captured by a plain vector representation.  
Many pairs of words have a clear entailment-relationship, for example:
\begin{center}
tiger $\Rightarrow$ big cat  $\Rightarrow$ mammal  $\Rightarrow$ vertebrate  $\Rightarrow$ animal   
\end{center}
While plain vectors living in a vector space do not come with any kind of structure that can capture these entailment-relationships, density operators can be partially ordered \cite{CoeckeMartin, bankova2016graded, vandeWeteringen}, and this partial order can then be interpreted as lexical entailment \cite{EsmaSC, BalkirKartsaklisSadrzadeh2015, bankova2016graded}.\footnote{In the first two of these papers, the ordering is taken to be a preorder for the sake of simplicity, with the induced equivalence classes corresponding to the lattice of closed subspaces, i.e~quantum logic \cite{BvN}. In the case of the partial orders of \cite{CoeckeMartin, bankova2016graded}, quantum logic embeds within the ordering of density operators.}   Since the space of density operators embeds in a vector space, we can rely on sums in order to construct general meanings from more specific ones e.g.:
\begin{center}
big cat := ${1\over N}$ (lion + tiger + cheetah + leopard + ...)    
\end{center}
 
 This brings up a dilemma: should we either use density operators to express ambiguity, or, lexical entailment?  We resolve this dilemma by introducing \em dual density operators\em.   These are mathematical entities which admit `two independent dimensions' of being a density operator. Moreover, just like ordinary density operators, these dual density operators inhabit a category of the appropriate structure for composing meanings, which is obtained by twice applying the CPM-construction. Hence they allow for data-driven and grammar-driven compositional reasoning about meanings of sentences, accounting for ambiguity as well as lexical entailment.    
 
 In the following section, we provide a direct construction of dual density operators, using standard Dirac notation.  In Section \ref{sec:catcons}, we provide the corresponding categorical construction.  Then, we provide an example encoding of meanings both involving ambiguity and lexical entailment, and in the following section we compose these meanings.  Finally, in Section \ref{sec:axiom}, we axiomatise categories resulting from twice applying the CPM-construction,  exposing  two contexts and two corresponding discarding operations. 
 
 \section{Direct construction}
  
 Given a set of normalised vectors $\{|\varphi_i\rangle\}$ in a finite-dimensional  inner-product  space $H$ and a probability distribution $\{p_i\}$ we  form a \em density operator for $H$ \em as follows:
 \[ 
\left( \{|\varphi_i\rangle\}, \{p_i\}\right)\ \  \mapsto \ \  \rho_{\rm operator} :=  \sum_i p_i | \varphi_i \rangle\langle \varphi_i |
 \] 
 That is, first we replace each vector by the pair consisting of the vector (a.k.a~`ket') and its adjoint  (a.k.a~`bra'), which together form a rank 1 operator. Then, we make a weighted sum.  Alternatively, instead of taking the adjoint of the vector, we could take its conjugate, and  instead of an operator obtain a two-system vector:      
 \beq\label{eq:dens-construction2}
\left( \{|\varphi_i\rangle\}, \{p_i\}\right)\ \  \mapsto \ \  |\rho\rangle :=  \sum_i p_i | \varphi_i \rangle\overline{|\varphi_i \rangle}  
 \eeq
One big advantage of \em density vectors for $H$ \em as compared to density operators, is that density vectors still live in a vector space $H \otimes \overline{H}$, where  $\overline{H}$ is the conjugate space,\footnote{For simplicity, one could take $H$ to be self-dual  so that $\overline{H}=H$.  However, some of the categorical constructions are directly guided by distinguishing between these two spaces.}  so that we can simply repeat construction (\ref{eq:dens-construction2}).  Doing so we obtain:
 \beq\label{eq:dens-construction3}
\left( \{|\rho_k\rangle\}, \{p_k'\}\right)\ \  \mapsto \ \  \sum_k p_k' | \rho_k \rangle\overline{|\rho_k \rangle}
\eeq
We will follow the  convention that conjugation of a state in $H \otimes H$ also swaps the states:
\[
\overline{|\rho_1 \rangle|\rho_2 \rangle} = \overline{|\rho_2 \rangle}\ \overline{|\rho_1 \rangle}
\] 
and hence chaining (\ref{eq:dens-construction2}) and (\ref{eq:dens-construction3}) together we obtain:  
 \beq\label{eq:dens-construction4}
\left( \{|\varphi_{ik}\rangle\}, \{p_{ik}\}, \{p_k'\}\right)\   \ \mapsto\  \  
\sum_k p_k' \left(\sum_i p_{ik} | \varphi_{ik} \rangle\overline{|\varphi_{ik} \rangle}\right)\overline{\left(\sum_j p_{jk} | \varphi_{jk} \rangle\overline{|\varphi_{jk} \rangle}\right)}
\eeq  
So, we obtain a vector in $H \otimes \overline{H} \otimes H \otimes \overline{H}$:
\beq\label{eq:dual-dens-form}
\Phi := \sum_{ijk} p_{ik} p_{jk} p_k' | \varphi_{ik} \rangle\overline{|\varphi_{ik} \rangle} | \varphi_{jk} \rangle\overline{|\varphi_{jk} \rangle}
\eeq      

As is obvious from the form in the RHS of (\ref{eq:dens-construction4}), the vector $\Phi$ can be seen as a density vector for $H \otimes  \overline{H}$.  However,  if we swap the 2nd and 4th vectors in (\ref{eq:dual-dens-form}), we obtain another density vector for $H \otimes  \overline{H}$:  
\beq\label{eq:dual-dens-formbis}
\sum_{ijk} p_{ik} p_{jk} p_k'  | \varphi_{ik} \rangle  \overline{|\varphi_{jk} \rangle} | \varphi_{jk} \rangle \overline{|\varphi_{ik} \rangle} 
=
\sum_{ijk} p_{ik} p_{jk} p_k' \left(|\varphi_{ik} \rangle \overline{| \varphi_{jk} \rangle}\right)\overline{\left(|\varphi_{ik} \rangle \overline{| \varphi_{jk}} \rangle\right)}
\eeq
Hence, the vector $\Phi$ can be thought of in two manners as a density vector for $H \otimes  \overline{H}$, and hence, can be thought of in two manners as a density operator for $H \otimes  \overline{H}$.  

We will refer to vectors in $H \otimes \overline{H} \otimes H \otimes \overline{H}$ of the form (\ref{eq:dual-dens-form}) as  \em Dual density operators for $H$\em.   Since to any dual density operator $\Phi$ correspond two density  vectors for $H \otimes  \overline{H}$: 
\[
\Phi_1:= \sum_{ijk} p_{ik} p_{jk} p_k' \left(|\varphi_{ik} \rangle \overline{| \varphi_{ik} \rangle}\right)\overline{\left(|\varphi_{jk} \rangle \overline{| \varphi_{jk}} \rangle\right)}
\qquad\quad
\Phi_2:= \sum_{ijk} p_{ik} p_{jk} p_k' \left(|\varphi_{ik} \rangle \overline{| \varphi_{jk} \rangle}\right)\overline{\left(|\varphi_{ik} \rangle \overline{| \varphi_{jk}} \rangle\right)}
\]
and hence two density operators for $H \otimes  \overline{H}$,  all features of density operators apply in two-fold to dual density operators.  For example, there are two notions of eigenvectors, two  notions of spectrum, two notions of entropy, two notions of (im)purity, and so on.
 
  
\section{Categorical construction}\label{sec:catcons}      

The direct construction of density vectors  from vectors is an instance of a  general category-theoretic construction, called the \em CPM-construction\em, which not only applies to inner-product spaces, but to any structure that can be organised in a so-called dagger compact closed category \cite{SelingerCPM}.  Moreover, in the case of inner-product spaces, it doesn't just generate density vectors in that case, but also completely positive maps.  In general, we again obtain a dagger compact closed category, so we can apply the CPM-construction as many times as we wish.

What this construction does is most easily seen in terms of the diagrammatic language of dagger compact closed categories \cite{SelingerSurvey}.\footnote{Please see \cite{CatsII} for a tutorial.}  In this language, inner-product spaces are represented by wires, and linear maps by boxes:     
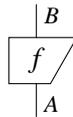
\ctikzfig{smapAB}
Vectors in $H$, when represented as linear maps from the vector space field $\mathbb{K}$ (seen as a one-dimensional inner-product space) into $H$, correspond to boxes without inputs, which in general we represent by triangles.  Conjugation is represented by horizontal reflection of these boxes, and we will make use of one special linear map with two inputs, and no outputs, i.e.~an \em effect\em, which we represent by a cap:  
\[
\raisebox{1mm}{\begin{tikzpicture}
	\begin{pgfonlayer}{nodelayer}
		\node [style=none] (0) at (-0.75, -0.5) {};
		\node [style=none] (1) at (0.75, -0.5) {};
		\node [style=none] (2) at (-0.75, -0.5) {}; 
		\node [style=none] (4) at (0.75, -0.5) {};
	\end{pgfonlayer}
	\begin{pgfonlayer}{edgelayer}
		\draw [in=90, out=90, looseness=1.75] (0.center) to (1.center); 
	\end{pgfonlayer}
\end{tikzpicture}}\ : H\otimes \overline{H} \to \mathbb{K}:: |\varphi\rangle\overline{|\varphi'\rangle}\mapsto \langle \varphi' | \varphi \rangle
\]

The CPM-construction  boils down to passing from general boxes to those of the form:          
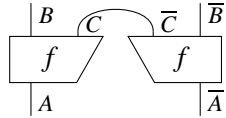
\ctikzfig{CPexer}  
When comparing this diagrams to the form (\ref{eq:dens-construction3}), the cup corresponds to the summation,  the type $C$ to the set of indices, and the probabilities are absorbed within the boxes. In fact, the vectors that we obtain in this manner are not normalised, and if we want to restrict to normalised ones, we require `trace preservation':
\ctikzfig{CPexernorm}

The \em CPM${}^2$-construction \em means applying the CPM-construction twice, yielding boxes of the form:
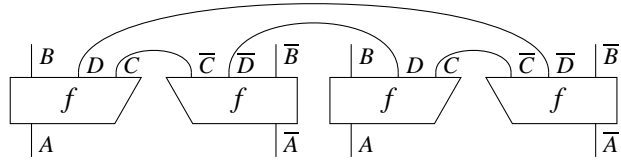
\ctikzfig{CPexer2}
and the  dual density operators $\Phi$ are then of the form:  
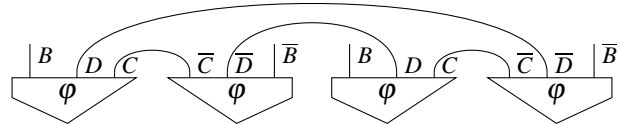
\ctikzfig{CPexer2state}
The density operator $\Phi_1$ is obtained by bending two wires down:  
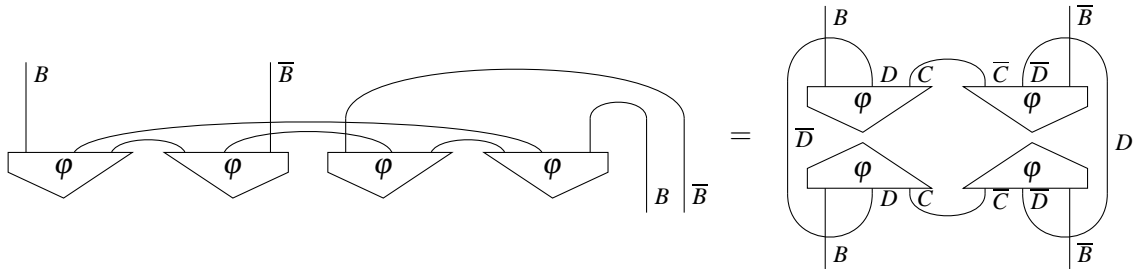
\ctikzfig{CPexer2statePhi1}
and the density operator $\Phi_2$ by doing the same after swapping the 1st and 3th wire:
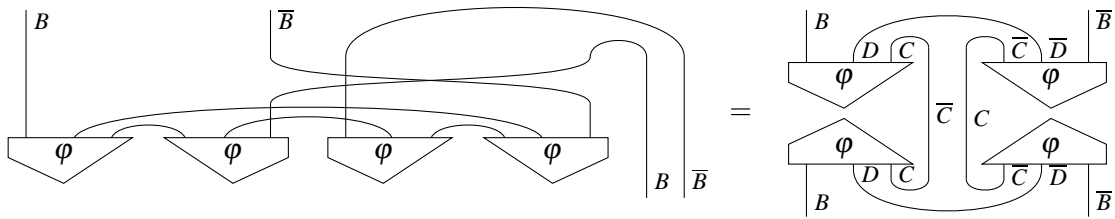
\ctikzfig{CPexer2statePhi2}

Note also that from the above it is  obvious that the two density operators $\Phi_1$ and $\Phi_2$  exist on `equal footing'.    More specifically, there is an isomorphism which takes the density operators of the form $\Phi_1$ to those of the form $\Phi_2$, which is realised by swapping the SW wire and the NE wire.   

Moreover, it also becomes clear that the two notions of (im)purity are independent, in the case of $\Phi_1$ depending on the `size' of $D$, while in the case of $\Phi_2$ it depends on the size of $C$, since it are wires of these respective types that connect the inputs and the outputs of the respective density operators.

\section{Ambiguity and lexical entailment}      

Dual density operators now provide a natural setting to accommodate both ambiguity and lexical entailment in natural language.  Given a dual density operator $\Phi$, the first density operator $\Phi_1$ accounts for entailment, while the dual structure, in addition, allows one to express ambiguity.  Theoretically, all meanings and their entailment relationships are encoded as density operators on $H$ and their partial ordering.  Here, all meanings are to be conceived as unambiguous, cf.~``queen'' as monarch and ``queen'' as rock band each have their own dedicated density operator.  Then, by construction (\ref{eq:dens-construction3}), we can introduce ambiguity.  For example, let ``Beirut" be the ambiguous  word with unambiguous meanings ``Beirut city" and ``Beirut band". The city of Beirut has neighbourhoods ``Ashrafieh", that we will denote by ``A", and ``Monot", that we will denote by ``M", while the band has members ``Zach", denoted by ``Z", and ``Paul", denoted by ``P".  We can use density operators:
\begin{center}
``Beirut city" $:= A\overline{A} + M\overline{M}$ \qquad\qquad\qquad\qquad ``Beirut band" $:= Z\overline{Z} + P\overline{P}$
\end{center}
in order to express that $A$ and $M$ entail ``Beirut city" and $Z$ and $P$ entail ``Beirut band", and we obtain the unambiguous meaning by first turning these in dual density operators and then adding them: 
\beqa
\mbox{\rm ``Beirut"} &:= & (A\overline{A} + M\overline{M})(A\overline{A} + M\overline{M})  + (Z\overline{Z} + P\overline{P})(Z\overline{Z} + P\overline{P}) \\
&:= &
A\overline{A}A\overline{A} + A\overline{A}M\overline{M} + M\overline{M}A\overline{A} + M\overline{M}M\overline{M}
+ Z\overline{Z}Z\overline{Z} + Z\overline{Z}P\overline{P} + P\overline{P}Z\overline{Z} + P\overline{P}P\overline{P}
\eeqa
Note that we did not add weights in order to keep the notation simple.

\begin{remark}
The procedure outlined above is not the only one for building meaning involving both ambiguity and lexical entailment.  An alternative one is presented in the first author's MSc thesis \cite{DanielaMSc}, which  relates  lexical entailment and  ambiguity directly to $\Phi_1$ and $\Phi_2$  respectively:  
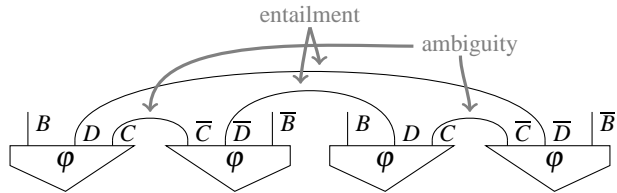
\ctikzfig{CPexer2state_copy}
The relationship between the alternative encodings is subject to currently ongoing research.
\end{remark}

\section{Interacting meanings}

In \cite{CSC}  a mathematical framework is proposed which allows for the computation of the meaning of sentences in terms of their constituents. This framework unifies two orthogonal but complementary models of meaning. 

The first  one formalises the grammar of natural language, for example, in terms of  \em pregroups \em  $(P,\leq,\cdot,1,(-)^{l},(-)^{r})$ where $(P,\leq,\cdot,1)$ is a partially ordered monoid, $(-)^{l}$ and $(-)^{r}$ are unary operations on $P$, called the left and right adjoints, satisfying the following inequalities for all $a \in P$:
\[
a^{l} \cdot a \leq 1 \leq a \cdot a^{l} \qquad\qquad\qquad\qquad a \cdot a^{r} \leq 1 \leq a^{r} \cdot a
\]
In what follows, we omit the ``$\cdot$" and replace ``$\leq$" by ``$\rightarrow$". To see how pregroups model grammar, we fix two basic grammatical types $\{n,s\}$, where \textit{n} is the grammatical type for \textit{noun}, and \textit{s} is the grammatical type for \textit{sentence}. Compound types are formed by adjoining and juxtaposing basic types: a transitive verb interacts with a subject to its left and an object to its right, to produce a sentence that is grammatically valid. Transitive verbs are therefore assigned the type $n^{r}sn^{l}$, and a transitive sentence reduces to a valid grammatical sentence as follows:
\[
n(n^{r}sn^{l})n = (nn^{r})s(n^{l}n) \rightarrow s
\]

The second approach concerns the distributional model of meaning, in which words are represented by vectors  in finite-dimensional inner-product spaces. While this model does not account for grammar, it does provide a reliable meaning for words. The algorithm of \cite{CSC} exploits the fact that pregroups on the one hand, when viewed as thin monoidal categories,  and inner-product spaces and linear maps on the other hand, are both examples of  compact-closed categories.  Then, via a strong monoidal functor between these two categories,  grammatical reductions are mapped on a linear map:
\[
[n(n^{r}sn^{l})n \rightarrow s] \quad\mapsto\quad \begin{tikzpicture}
	\begin{pgfonlayer}{nodelayer}
		\node [style=none] (0) at (0, -0.5) {};  
		\node [style=none] (1) at (0, 0.5) {};
		\node [style=none] (2) at (-2.5, -0.5) {};
		\node [style=none] (3) at (-1, -0.5) {};
		\node [style=none] (4) at (-2.5, -0.5) {};
		\node [style=none] (5) at (1, -0.5) {};
		\node [style=none] (6) at (2.5, -0.5) {};
		\node [style=none] (7) at (1, -0.5) {};
	\end{pgfonlayer}
	\begin{pgfonlayer}{edgelayer}
		\draw [style=swap] (0.center) to (1.center);
		\draw [style=swap, in=90, out=90, looseness=1.50] (4.center) to (3.center);
		\draw [style=swap, in=90, out=90, looseness=1.50] (5.center) to (6.center);  
	\end{pgfonlayer}
\end{tikzpicture}  
\]
which then when applied to meaning vectors, gives the meaning of a sentence:     
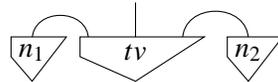
\ctikzfig{typered-copy}

Clearly, the use of a category of inner-product spaces and linear maps is not at all essential; it suffices to have any compact-closed category, or even more general, a category that matches the structure of the chosen categorial grammar \cite{LambekvsLambek}. Since the CPM-construction maps a dagger compact closed category on a dagger compact closed category \cite{SelingerCPM}, rather than using vectors, we can use density operators to represent meanings, or, of course, dual density operators.

To illustrate this, let us go back to our example involving Beirut. We seek to show that the meaning of ambiguous words `collapses' when enough context is provided. For this, we will compute the meanings of two noun phrases:  ``Beirut that plays at Beirut", and ``Beirut that Beirut plays at". We expect the former to be ``Beirut band",
and the latter to be ``Beirut city". We already gave the meaning of ``Beirut", so it suffices to give the meaning of ``play-at''.  It is a transitive verb which we take  to be non-ambiguous, and atomic.  Hence, in essence it is described by a vector in $N\otimes S \otimes N$ where $N$ is the space in which we describe nouns, namely the one we used to construct ``Beirut", and $S$ is the sentence space, which for the sake of simplicity we choose to be $\{\bot, \top\}$, where $\bot$ stands for ``false" and $\top$ for ``true".   A natural way for constructing the meaning of a verb, is to simply take pairs of objects and subjects which `obey' that verb, with a ``true"-symbol in the middle.  Therefore, for ``play-at'' as a vector in $N\otimes S \otimes N$ we set:
\[
\mbox{\rm{play-at}}_{N\otimes S \otimes N} := Z\top A + P\top A
\]
meaning that Zach and Paul play in neighbourhood Ashrafieh. As a dual density operator this gives:  
\[
\mbox{\rm{``play-at''}} := (Z\top A + P\top A)(\overline{Z}\top \overline{A} + \overline{P}\top \overline{A})(Z\top A + P\top A)(\overline{Z}\top \overline{A} + \overline{P}\top \overline{A})  
\]

We follow  \cite{FrobMeanI} in order to assign meaning to the relative pronoun ``that''.  Diagrammatically, this boils down to the use of `spiders', and category-theoretically, the use of  special commutative Frobenius algebras.  Given an ONB we will make use of:
\[
\raisebox{1mm}{\begin{tikzpicture}
	\begin{pgfonlayer}{nodelayer}
		\node [style=white dot] (0) at (0, -0.5) {};
		\node [style=none] (1) at (0, 0.5) {};
		\node [style=none] (2) at (1, 0.5) {};
		\node [style=none] (3) at (-1, 0.5) {};
	\end{pgfonlayer}
	\begin{pgfonlayer}{edgelayer}
		\draw [style=swap] (0) to (1.center);
		\draw [style=none, bend right=45, looseness=1.00] (3.center) to (0);
		\draw [style=none, bend left=45, looseness=1.00] (2.center) to (0);
	\end{pgfonlayer}
\end{tikzpicture}}\ : \mathbb{K} \to H\otimes H\otimes H :: 1\mapsto \sum_i | i ii \rangle
\qquad\qquad\qquad
\raisebox{1mm}{\begin{tikzpicture}
	\begin{pgfonlayer}{nodelayer}
		\node [style=white dot] (0) at (0, -0.5) {};
		\node [style=none] (1) at (0, 0.5) {};
	\end{pgfonlayer}
	\begin{pgfonlayer}{edgelayer}
		\draw [style=swap] (0) to (1.center);
	\end{pgfonlayer}
\end{tikzpicture}}\ : \mathbb{K} \to H :: 1\mapsto \sum_i | i  \rangle  
\]
The grammatical type of ``that''  used as a subject relative pronoun is $N\otimes N\otimes S \otimes N$, while as an object relative pronoun it is $N\otimes N \otimes N\otimes S$, and we set:
\[
\mbox{\rm{``that"}}_{subj} :=\ \  \begin{tikzpicture}
	\begin{pgfonlayer}{nodelayer}
		\node [style=white dot] (0) at (0, -0.75) {};
		\node [style=none] (1) at (-0.75, 1) {};
		\node [style=none] (2) at (2.25, 1) {};
		\node [style=none] (3) at (-2.25, 1) {};
		\node [style=none] (4) at (0.75, 1) {};
		\node [style=white dot] (5) at (0.75, 0) {};
		\node [style=right label, yshift=0.5mm] (6) at (2.5, 0.5) {$N$};
		\node [style=right label, yshift=0.5mm] (7) at (-0.5, 0.5) {$N$};
		\node [style=right label, yshift=0.5mm] (8) at (-2, 0.5) {$N$};
		\node [style=right label, yshift=0.5mm] (9) at (1, 0.5) {$S$};
	\end{pgfonlayer}
	\begin{pgfonlayer}{edgelayer}
		\draw [style=swap, in=-90, out=120, looseness=1.00] (0) to (1.center);
		\draw [style=none, in=180, out=-90, looseness=1.00] (3.center) to (0);
		\draw [style=none, in=0, out=-90, looseness=1.00] (2.center) to (0);
		\draw [style=swap] (5) to (4.center);
	\end{pgfonlayer}
\end{tikzpicture}    
\qquad\qquad\qquad
\mbox{\rm{``that"}}_{obj} :=\ \  
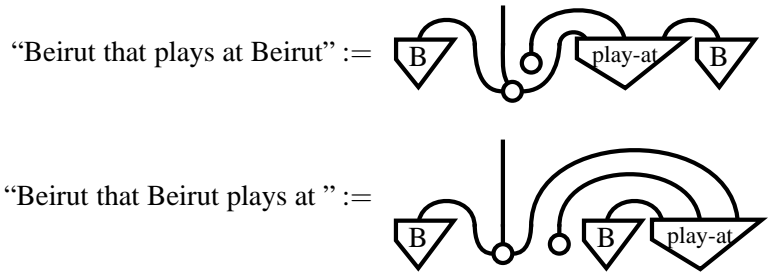
\begin{tikzpicture}
	\begin{pgfonlayer}{nodelayer}
		\node [style=white dot] (0) at (-0.75, -0.75) {};
		\node [style=none] (1) at (-0.75, 1) {};
		\node [style=none] (2) at (0.75, 1) {};
		\node [style=none] (3) at (-2.25, 1) {};
		\node [style=none] (4) at (2.25, 1) {};
		\node [style=white dot] (5) at (2.25, 0) {};
		\node [style=right label, yshift=0.5mm] (6) at (1, 0.5) {$N$};
		\node [style=right label, yshift=0.5mm] (7) at (-0.5, 0.5) {$N$};
		\node [style=right label, yshift=0.5mm] (8) at (-2, 0.5) {$N$};
		\node [style=right label, yshift=0.5mm] (9) at (2.5, 0.5) {$S$};
	\end{pgfonlayer}
	\begin{pgfonlayer}{edgelayer}
		\draw [style=swap, in=-90, out=90, looseness=1.00] (0) to (1.center);
		\draw [style=none, in=180, out=-90, looseness=1.00] (3.center) to (0);
		\draw [style=none, in=0, out=-90, looseness=1.00] (2.center) to (0);
		\draw [style=swap] (5) to (4.center);
	\end{pgfonlayer}
\end{tikzpicture}  
\]
So:
\[
\mbox{\rm{``Beirut that plays at Beirut"}} :=\ \  \begin{tikzpicture}
	\begin{pgfonlayer}{nodelayer}
		\node [style=white ddot] (0) at (-1.25, -1) {};
		\node [style=none] (1) at (-1.5, 0.25) {};
		\node [style=none] (2) at (0, 0.25) {};
		\node [style=none] (3) at (-2.25, 0.25) {};
		\node [style=none] (4) at (-0.75, 0.25) {};
		\node [style=white ddot] (5) at (-0.75, -0.25) {};
		\node [style=none] (6) at (0, 0.25) {};
		\node [style=none] (7) at (0.75, 0.25) {};
		\node [style=none] (8) at (4.25, 0.25) {};
		\node [style=dkpoint, yshift=-0.4mm] (9) at (4.25, 0) {B};
		\node [style=none] (10) at (1.75, 0.25) {};
		\node [style=none] (11) at (2.75, 0.25) {};
		\node [style=none] (12) at (0, 0.25) {};
		\node [style=none] (13) at (2.75, 0.25) {};
		\node [style=wide dkpoint, yshift=-0.4mm] (14) at (1.75, 0) {\footnotesize play-at};
		\node [style=none] (15) at (-0.75, 0.25) {};
		\node [style=none] (16) at (-2.25, 0.25) {};
		\node [style=dkpoint, yshift=-0.4mm] (17) at (-3.75, 0) {B};
		\node [style=none] (18) at (-3.75, 0.25) {};
		\node [style=none] (19) at (-2.25, 0.25) {};
		\node [style=none] (20) at (-1.5, 0.25) {};
		\node [style=none] (21) at (-1.5, 1.25) {};
	\end{pgfonlayer}
	\begin{pgfonlayer}{edgelayer}
		\draw [style=boldedge, in=-90, out=120, looseness=0.75] (0) to (1.center);
		\draw [style=boldedge, in=180, out=-90, looseness=1.00] (3.center) to (0);
		\draw [style=boldedge, in=0, out=-90, looseness=1.00] (2.center) to (0);
		\draw [style=boldedge] (5) to (4.center);
		\draw [style=boldedge, bend right=90, looseness=1.25] (10.center) to (15.center);
		\draw [style=boldedge, in=90, out=90, looseness=1.50] (12.center) to (7.center);
		\draw [style=boldedge, in=90, out=90, looseness=1.50] (11.center) to (8.center);
		\draw [style=boldedge, in=90, out=90, looseness=1.50] (16.center) to (18.center);
		\draw [style=boldedge] (20.center) to (21.center);
	\end{pgfonlayer}
\end{tikzpicture}
\]
\[
\mbox{\rm{``Beirut that Beirut plays at "}} :=\ \  \begin{tikzpicture}
	\begin{pgfonlayer}{nodelayer}
		\node [style=white ddot] (0) at (-0.75, -1.5) {};
		\node [style=none] (1) at (-0.75, -0.75) {};
		\node [style=none] (2) at (0, -0.75) {};
		\node [style=none] (3) at (-1.5, -0.75) {};
		\node [style=none] (4) at (0.75, -0.75) {};
		\node [style=white ddot] (5) at (0.75, -1.25) {};
		\node [style=none] (6) at (0, -0.75) {};
		\node [style=none] (7) at (5.5, -0.75) {};
		\node [style=none] (8) at (2, -0.75) {};
		\node [style=dkpoint, yshift=-0.4mm] (9) at (2, -1) {B};
		\node [style=none] (10) at (4.5, -0.75) {};
		\node [style=none] (11) at (3.5, -0.75) {};
		\node [style=none] (12) at (0, -0.75) {};
		\node [style=none] (13) at (3.5, -0.75) {};
		\node [style=wide dkpoint, yshift=-0.4mm] (14) at (4.5, -1) {\footnotesize play-at};
		\node [style=none] (15) at (0.75, -0.75) {};
		\node [style=none] (16) at (-1.5, -0.75) {};
		\node [style=dkpoint, yshift=-0.4mm] (17) at (-3, -1) {B};
		\node [style=none] (18) at (-3, -0.75) {};
		\node [style=none] (19) at (-1.5, -0.75) {};
		\node [style=none] (20) at (-0.75, -0.75) {};
		\node [style=none] (21) at (-0.75, 1.5) {};
	\end{pgfonlayer}
	\begin{pgfonlayer}{edgelayer}
		\draw [style=boldedge, in=-90, out=90, looseness=0.75] (0) to (1.center);
		\draw [style=boldedge, in=180, out=-90, looseness=1.00] (3.center) to (0);
		\draw [style=boldedge, in=0, out=-90, looseness=1.00] (2.center) to (0);
		\draw [style=boldedge] (5) to (4.center);
		\draw [style=boldedge, bend right=90, looseness=1.25] (10.center) to (15.center);
		\draw [style=boldedge, in=90, out=90, looseness=1.25] (12.center) to (7.center);
		\draw [style=boldedge, in=90, out=90, looseness=1.50] (11.center) to (8.center);
		\draw [style=boldedge, in=90, out=90, looseness=1.50] (16.center) to (18.center);
		\draw [style=boldedge] (20.center) to (21.center);
	\end{pgfonlayer}
\end{tikzpicture}
\]
where the use of bold-wires indicates that all meanings are dual density operators.  A somewhat tedious direct computation of these diagrams then indeed yields:
\[
\mbox{\rm{``Beirut that plays at Beirut"}} := \mbox{\rm{``Beirut-band"}}
\qquad\quad
\mbox{\rm{``Beirut that Beirut plays at "}} := \mbox{\rm{``Beirut-city"}}
\]
Both results are consistent with our expectations and accurately model the case where enough context is provided to disambiguate the meaning of a word. Further examples are provided in \cite{DanielaMSc}.

\section{Axiomatic characterisation}\label{sec:axiom}  

Density operators allow for discarding part of the description of a composite system, where the discarded part corresponds to the environment or context. As shown in \cite{SelingerAxiom, CPer}, the CPM-construction can be recast in terms of  an \em environment structure \em on a dagger compact closed category $\textbf{C}$, which consists   of a designated effect $\top_{A}: A \rightarrow I$ for each object $A$ in $\textbf{C}$, called \textit{discarding}, obeying $\top_{I} = 1_{I}$, $\top_{A \otimes B}=\top_{A} \otimes \top_{B}$, and $({\top_{A}})_{*}= \top_{A^{*}}$, together with an all-objects-including sub-dagger compact closed category $\textbf{C}_{\Sigma}$ of \textit{pure morphisms}, which is such that for all pure morphisms $f, g$ we have:
\begin{equation}  
\label{axiom1}
\begin{tikzpicture}
	\begin{pgfonlayer}{nodelayer}
		\node [style=map] (0) at (-7, -1) {$f$};
		\node [style=none] (1) at (-7, -2.25) {};
		\node [style=none] (2) at (-7, -1.5) {};
		\node [style=none] (3) at (-7, -0.5) {};
		\node [style=none] (4) at (-7, -0.5) {};
		\node [style=none] (5) at (-5, 0) {$=$};
		\node [style=mapadj] (6) at (-7, 1) {$f$};
		\node [style=none] (7) at (-7, 2.25) {};
		\node [style=none] (8) at (-7, 1.5) {};
		\node [style=none] (9) at (-7, 0.5) {};
		\node [style=none] (10) at (-7, 0.5) {};
		\node [style=map] (11) at (-3.25, -1) {$g$};
		\node [style=none] (12) at (-3.25, 0.5) {};
		\node [style=none] (13) at (-3.25, -1.5) {};
		\node [style=none] (14) at (-3.25, -0.5) {};
		\node [style=none] (15) at (-3.25, 0.5) {};
		\node [style=none] (16) at (-3.25, 2.25) {};
		\node [style=none] (17) at (-3.25, -2.25) {};
		\node [style=none] (18) at (-3.25, 1.5) {};
		\node [style=none] (19) at (-3.25, -0.5) {};
		\node [style=mapadj] (20) at (-3.25, 1) {$g$};
		\node [style=none] (21) at (-0.25, 0) {$\Longleftrightarrow$};
		\node [style=none] (22) at (2.5, -0.5) {};
		\node [style=none] (23) at (2.5, -1.25) {};
		\node [style=map] (24) at (2.5, 0) {$f$};
		\node [style=none] (25) at (2.5, 1.25) {};
		\node [style=none] (26) at (2.5, 0.5) {};
		\node [style=copoint, yshift=-0.4mm] (27) at (2.5, 1.5) {$\top$};
		\node [style=none] (28) at (6.25, -1.25) {};
		\node [style=none] (29) at (6.25, 1.25) {};
		\node [style=copoint, yshift=-0.4mm] (30) at (6.25, 1.5) {$\top$};
		\node [style=map] (31) at (6.25, 0) {$g$};
		\node [style=none] (32) at (6.25, -0.5) {};
		\node [style=none] (33) at (6.25, 0.5) {};
		\node [style=none] (34) at (4.5, 0) {$=$};
	\end{pgfonlayer}
	\begin{pgfonlayer}{edgelayer}
		\draw [style=swap] (1.center) to (2.center);
		\draw [style=swap] (7.center) to (8.center);
		\draw [style=none] (9.center) to (4.center);
		\draw [style=swap] (17.center) to (13.center);
		\draw [style=swap] (16.center) to (18.center);
		\draw [style=none] (12.center) to (19.center);
		\draw [style=swap] (23.center) to (22.center);
		\draw [style=swap] (25.center) to (26.center);
		\draw [style=swap] (28.center) to (32.center);
		\draw [style=swap] (29.center) to (33.center);
	\end{pgfonlayer}
\end{tikzpicture}
\end{equation}  
Applying  (\ref{axiom1}) to the specific case of vectors yields:
\[
|\psi\rangle\langle  \psi |\ =\  |\varphi\rangle\langle\varphi | \ \ \Longleftrightarrow \ \ |\psi\rangle\ =\  |\varphi\rangle 
\]
which has been called  \textit{preparation-state agreement} \cite{SelingerAxiom}. In can then be shown that a dagger compact closed category $\textbf{C}$ carrying an environment structure is isomorphic to $\textbf{CPM}(\textbf{C}_{\Sigma})$, and applying the CPM-construction to a dagger compact closed category $\textbf{C}$ which satisfies  preparation-state agreement  induces an environment structure on $\textbf{C}$ \cite{SelingerAxiom, CPer}.    

Similarly,  a dual-environment structure on a dagger compact closed category $\textbf{C}$ consists of two discarding effects $\top_{1,A}, \top_{2,A}: A \rightarrow I$ for each object $A$ of $\textbf{C}$,  together with  an all-objects-including sub-dagger compact closed category $\textbf{C}_{\Sigma^{2}}$ of \textit{pure morphisms},   which is such that for all pure morphisms $f, g$ we have:
\ctikzfig{Axiom2}
Now, a dagger compact closed category $\textbf{C}$ carrying a dual-environment structure is isomorphic to 
$\textbf{CPM}^{2}$ $(\textbf{C}_{\Sigma^{2}})$, and applying the CPM$^{2}$-construction to a dagger compact closed category $\textbf{C}$ which satisfies the preparation-state agreement axiom induces a dual-environment structure on $\textbf{C}$.     

The proof of this fact can be found in \cite{DanielaMSc}, as well as a generalization to multiple applications of the CPM-construction, resulting in multiple discarding operations.

\section{Discussion and outlook} 

Firstly, we applied the CPM-construction twice, in order to accommodate two linguistic features, but there is no reason to stop there: more applications would enable one to accommodate more natural language features.  

Secondly, the same `trick' does not only apply to vectors in inner-product spaces, but any candidate model of meaning that can be structured in a dagger compact closed category.  One example of other models currently being studied in \cite{ConcSpac} are based on G\"ardenfors' conceptual spaces.  \cite{gardenfors}.

Thirdly, density operators were borrowed  from physics in order to represent ambiguity, perfectly matching their quantum-theoretical interpretation in terms of lack of knowledge.  When providing them with a partial ordering in order to represent lexical entailment, one actually went beyond the standard practice in physics, although a subset of the ordering is Birkhoff-von Neumann quantum logic.  However, dual density operators are an entirely new kind of mathematical entity that (to our knowledge) have never been used in physics.  This of course does not exclude that there is a natural application for them.    

Fourthly, of course, we only provided one very simple proof-of-concept example in support of our claims.  More involved examples as well as empirical evidence are needed to firmly establish dual density operators as a useful tool for representing natural language meaning.

Finally, many books have been written on density operators.  Several things that don't  make sense for vectors, emerge for density operators, like diagonalisability, spectrum, entropy and so on.  Dual density operators are yet again a new entity, and hence new basic mathematics needs to be developed.     

For example, we know that construction (\ref{eq:dens-construction2}) and application of the CPM-construction to inner-product spaces yields the same result.  However,  this isn't entirely true anymore for construction (\ref{eq:dens-construction4}) and twice applying the CPM-construction to inner-product spaces.  Indeed, in ongoing work in collaboration with Maaike Zwart we have characterised the dual density operators obtained via (\ref{eq:dens-construction4}) as a proper subset of those that arise from  twice applying the CPM-construction.  This is only the beginning, and much  more remains to be   discovered, for which we refer to a future publication.     

%

\bibliographystyle{eptcs}
\providecommand{\urlalt}[2]{\href{#1}{#2}}
\bibliography{main}

\begin{thebibliography}{10}
\providecommand{\bibitemdeclare}[2]{}
\providecommand{\surnamestart}{}
\providecommand{\surnameend}{}
\providecommand{\urlprefix}{Available at }
\providecommand{\url}[1]{\texttt{#1}}
\providecommand{\href}[2]{\texttt{#2}}
\providecommand{\urlalt}[2]{\href{#1}{#2}}
\providecommand{\doi}[1]{doi:\urlalt{http://dx.doi.org/#1}{#1}}
\providecommand{\bibinfo}[2]{#2}

\bibitemdeclare{mastersthesis}{DanielaMSc}
\bibitem{DanielaMSc}
\bibinfo{author}{D.~\surnamestart Ashoush\surnameend} (\bibinfo{year}{2015}):
  \emph{\bibinfo{title}{Categorical Models of Meaning: Accommodating for
  Lexical Ambiguity and Entailment}}.
\newblock Master's thesis, \bibinfo{school}{University of Oxford}.

\bibitemdeclare{misc}{BalkirKartsaklisSadrzadeh2015}
\bibitem{BalkirKartsaklisSadrzadeh2015}
\bibinfo{author}{E.~\surnamestart Balk{\i}r\surnameend},
  \bibinfo{author}{D.~\surnamestart Kartsaklis\surnameend} \&
  \bibinfo{author}{M.~\surnamestart Sadrzadeh\surnameend}
  (\bibinfo{year}{2015}): \emph{\bibinfo{title}{Sentence Entailment in
  Compositional Distributional Semantics}}.
\newblock \bibinfo{note}{International Symposium of Artificial Intelligence and
  Mathematics, 2016}.

\bibitemdeclare{incollection}{EsmaSC}
\bibitem{EsmaSC}
\bibinfo{author}{E.~\surnamestart Balkir\surnameend},
  \bibinfo{author}{M.~\surnamestart Sadrzadeh\surnameend} \&
  \bibinfo{author}{B.~\surnamestart Coecke\surnameend} (\bibinfo{year}{2016}):
  \emph{\bibinfo{title}{Distributional sentence entailment using density
  matrices}}.
\newblock In: {\sl \bibinfo{booktitle}{Topics in Theoretical Computer
  Science}}, \bibinfo{publisher}{Springer}, pp. \bibinfo{pages}{1--22},
  \doi{10.1007/978-3-319-28678-5}.

\bibitemdeclare{article}{bankova2016graded}
\bibitem{bankova2016graded}
\bibinfo{author}{D.~\surnamestart Bankova\surnameend},
  \bibinfo{author}{B.~\surnamestart Coecke\surnameend},
  \bibinfo{author}{M.~\surnamestart Lewis\surnameend} \&
  \bibinfo{author}{D.~\surnamestart Marsden\surnameend} (\bibinfo{year}{2016}):
  \emph{\bibinfo{title}{Graded Entailment for Compositional Distributional
  Semantics}}.
\newblock {\sl \bibinfo{journal}{\lnk{arXiv:1601.04908}}}.

\bibitemdeclare{article}{BvN}
\bibitem{BvN}
\bibinfo{author}{G.~\surnamestart Birkhoff\surnameend} \&
  \bibinfo{author}{J.~\surnamestart von Neumann\surnameend}
  (\bibinfo{year}{1936}): \emph{\bibinfo{title}{The logic of quantum
  mechanics}}.
\newblock {\sl \bibinfo{journal}{Annals of Mathematics}} \bibinfo{volume}{37},
  pp. \bibinfo{pages}{823--843}, \doi{10.2307/1968621}.

\bibitemdeclare{inproceedings}{ConcSpac}
\bibitem{ConcSpac}
\bibinfo{author}{J.~\surnamestart Bolt\surnameend},
  \bibinfo{author}{B.~\surnamestart Coecke\surnameend},
  \bibinfo{author}{F.~\surnamestart Genovese\surnameend},
  \bibinfo{author}{M.~\surnamestart Lewis\surnameend},
  \bibinfo{author}{D.~\surnamestart Marsden\surnameend} \&
  \bibinfo{author}{R.~\surnamestart Piedeleu\surnameend}
  (\bibinfo{year}{2016}): \emph{\bibinfo{title}{Interacting Conceptual
  Spaces}}.
\newblock In: {\sl \bibinfo{booktitle}{Semantic Spaces at the Intersection of
  NLP, Physics and Cognitive Science}}.

\bibitemdeclare{article}{teleling}
\bibitem{teleling}
\bibinfo{author}{S.~\surnamestart Clark\surnameend},
  \bibinfo{author}{B.~\surnamestart Coecke\surnameend},
  \bibinfo{author}{E.~\surnamestart Grefenstette\surnameend},
  \bibinfo{author}{S.~\surnamestart Pulman\surnameend} \&
  \bibinfo{author}{M.~\surnamestart Sadrzadeh\surnameend}
  (\bibinfo{year}{2014}): \emph{\bibinfo{title}{A quantum teleportation
  inspired algorithm produces sentence meaning from word meaning and
  grammatical structure}}.
\newblock {\sl \bibinfo{journal}{Malaysian Journal of Mathematical Sciences}}
  \bibinfo{volume}{8}, pp. \bibinfo{pages}{15--25}.
\newblock \bibinfo{note}{\lnk{arXiv:1305.0556}}.

\bibitemdeclare{article}{SelingerAxiom}
\bibitem{SelingerAxiom}
\bibinfo{author}{B.~\surnamestart Coecke\surnameend} (\bibinfo{year}{2008}):
  \emph{\bibinfo{title}{Axiomatic description of mixed states from {S}elinger's
  {CPM}-construction}}.
\newblock {\sl \bibinfo{journal}{Electronic Notes in Theoretical Computer
  Science}} \bibinfo{volume}{210}, pp. \bibinfo{pages}{3--13},
  \doi{10.1016/j.entcs.2008.04.014}.

\bibitemdeclare{article}{LambekvsLambek}
\bibitem{LambekvsLambek}
\bibinfo{author}{B.~\surnamestart Coecke\surnameend},
  \bibinfo{author}{E.~\surnamestart Grefenstette\surnameend} \&
  \bibinfo{author}{M.~\surnamestart Sadrzadeh\surnameend}
  (\bibinfo{year}{2013}): \emph{\bibinfo{title}{Lambek vs. {L}ambek: Functorial
  vector space semantics and string diagrams for {L}ambek calculus}}.
\newblock {\sl \bibinfo{journal}{Annals of Pure and Applied Logic}}
  \bibinfo{volume}{164}, pp. \bibinfo{pages}{1079--1100},
  \doi{10.1016/j.apal.2013.05.009}.

\bibitemdeclare{incollection}{CoeckeMartin}
\bibitem{CoeckeMartin}
\bibinfo{author}{B.~\surnamestart Coecke\surnameend} \&
  \bibinfo{author}{K.~\surnamestart Martin\surnameend} (\bibinfo{year}{2011}):
  \emph{\bibinfo{title}{A partial order on classical and quantum states}}.
\newblock In \bibinfo{editor}{B.~\surnamestart Coecke\surnameend}, editor: {\sl
  \bibinfo{booktitle}{New Structures for Physics}}, \bibinfo{series}{Lecture
  Notes in Physics}, \bibinfo{publisher}{Springer}, pp.
  \bibinfo{pages}{593--683}, \doi{10.1007/978-3-642-12821-9}.

\bibitemdeclare{incollection}{CatsII}
\bibitem{CatsII}
\bibinfo{author}{B.~\surnamestart Coecke\surnameend} \&
  \bibinfo{author}{{\'E}.~O. \surnamestart Paquette\surnameend}: In
  \bibinfo{editor}{B.~\surnamestart Coecke\surnameend}, editor: {\sl
  \bibinfo{booktitle}{New Structures for Physics}},
  \doi{10.1007/978-3-642-12821-9}.

\bibitemdeclare{inproceedings}{CPer}
\bibitem{CPer}
\bibinfo{author}{B.~\surnamestart Coecke\surnameend} \&
  \bibinfo{author}{S.~\surnamestart Perdrix\surnameend} (\bibinfo{year}{2010}):
  \emph{\bibinfo{title}{Environment and classical channels in categorical
  quantum mechanics}}.
\newblock In: {\sl \bibinfo{booktitle}{Proceedings of the 19th EACSL Annual
  Conference on Computer Science Logic (CSL)}}, {\sl \bibinfo{series}{Lecture
  Notes in Computer Science}} \bibinfo{volume}{6247}, pp.
  \bibinfo{pages}{230--244}, \doi{10.1007/978-3-642-15205-4}.

\bibitemdeclare{incollection}{CSC}
\bibitem{CSC}
\bibinfo{author}{B.~\surnamestart Coecke\surnameend},
  \bibinfo{author}{M.~\surnamestart Sadrzadeh\surnameend} \&
  \bibinfo{author}{S.~\surnamestart Clark\surnameend} (\bibinfo{year}{2010}):
  \emph{\bibinfo{title}{Mathematical foundations for a compositional
  distributional model of meaning}}.
\newblock In \bibinfo{editor}{J.~\surnamestart van Benthem\surnameend},
  \bibinfo{editor}{M.~\surnamestart Moortgat\surnameend} \&
  \bibinfo{editor}{W.~\surnamestart Buszkowski\surnameend}, editors: {\sl
  \bibinfo{booktitle}{A Festschrift for Jim Lambek}}, {\sl
  \bibinfo{series}{Linguistic Analysis}}~\bibinfo{volume}{36}, pp.
  \bibinfo{pages}{345--384}.
\newblock \bibinfo{note}{\lnk{arXiv:1003.4394}}.

\bibitemdeclare{book}{gardenfors}
\bibitem{gardenfors}
\bibinfo{author}{Peter \surnamestart G{\"a}rdenfors\surnameend}
  (\bibinfo{year}{2014}): \emph{\bibinfo{title}{The Geometry of Meaning:
  Semantics Based on Conceptual Spaces}}.
\newblock \bibinfo{publisher}{MIT Press}.

\bibitemdeclare{article}{Gleason}
\bibitem{Gleason}
\bibinfo{author}{A.~M. \surnamestart Gleason\surnameend}
  (\bibinfo{year}{1957}): \emph{\bibinfo{title}{Measures on the closed
  subspaces of a {H}ilbert space}}.
\newblock {\sl \bibinfo{journal}{Journal of Mathematics and Mechanics}}
  \bibinfo{volume}{6}, pp. \bibinfo{pages}{885--893},
  \doi{10.1512/iumj.1957.6.56050}.

\bibitemdeclare{inproceedings}{GrefSadr}
\bibitem{GrefSadr}
\bibinfo{author}{E.~\surnamestart Grefenstette\surnameend} \&
  \bibinfo{author}{M.~\surnamestart Sadrzadeh\surnameend}
  (\bibinfo{year}{2011}): \emph{\bibinfo{title}{Experimental Support for a
  Categorical Compositional Distributional Model of Meaning}}.
\newblock In: {\sl \bibinfo{booktitle}{The 2014 Conference on Empirical Methods
  on Natural Language Processing.}}, pp. \bibinfo{pages}{1394--1404}.
\newblock \bibinfo{note}{\lnk{arXiv:1106.4058}}.

\bibitemdeclare{phdthesis}{DimitriDPhil}
\bibitem{DimitriDPhil}
\bibinfo{author}{D.~\surnamestart Kartsaklis\surnameend}
  (\bibinfo{year}{2014}): \emph{\bibinfo{title}{Compositional Distributional
  Semantics with Compact Closed Categories and Frobenius Algebras}}.
\newblock Ph.D. thesis, \bibinfo{school}{University of Oxford}.

\bibitemdeclare{inproceedings}{KartSadr}
\bibitem{KartSadr}
\bibinfo{author}{D.~\surnamestart Kartsaklis\surnameend} \&
  \bibinfo{author}{M.~\surnamestart Sadrzadeh\surnameend}
  (\bibinfo{year}{2013}): \emph{\bibinfo{title}{Prior disambiguation of word
  tensors for constructing Sentence vectors}}.
\newblock In: {\sl \bibinfo{booktitle}{The 2013 Conference on Empirical Methods
  on Natural Language Processing.}}, \bibinfo{publisher}{ACL}, pp.
  \bibinfo{pages}{1590--1601}.

\bibitemdeclare{article}{lundburgess1996}
\bibitem{lundburgess1996}
\bibinfo{author}{K.~\surnamestart Lund\surnameend} \&
  \bibinfo{author}{C.~\surnamestart Burgess\surnameend} (\bibinfo{year}{1996}):
  \emph{\bibinfo{title}{Producing high-dimensional semantic spaces from lexical
  co-occurrence.}}
\newblock {\sl \bibinfo{journal}{Behavior Research Methods, Instruments {\&}
  Computers}} \bibinfo{volume}{28}, pp. \bibinfo{pages}{203--208},
  \doi{10.3758/BF03204766}.

\bibitemdeclare{book}{vN}
\bibitem{vN}
\bibinfo{author}{J.~\surnamestart von Neumann\surnameend}
  (\bibinfo{year}{1932}): \emph{\bibinfo{title}{Mathematische grundlagen der
  quantenmechanik}}.
\newblock \bibinfo{publisher}{Springer-Verlag}.
\newblock \bibinfo{note}{Translation, {\it Mathematical foundations of quantum
  mechanics}, Princeton University Press, 1955.}

\bibitemdeclare{mastersthesis}{RobinMSc}
\bibitem{RobinMSc}
\bibinfo{author}{R.~\surnamestart Piedeleu\surnameend} (\bibinfo{year}{2014}):
  \emph{\bibinfo{title}{Ambiguity in Categorical Models of Meaning}}.
\newblock Master's thesis, \bibinfo{school}{University of Oxford}.

\bibitemdeclare{inproceedings}{calco2015}
\bibitem{calco2015}
\bibinfo{author}{Robin \surnamestart Piedeleu\surnameend},
  \bibinfo{author}{Dimitri \surnamestart Kartsaklis\surnameend},
  \bibinfo{author}{Bob \surnamestart Coecke\surnameend} \&
  \bibinfo{author}{Mehrnoosh \surnamestart Sadrzadeh\surnameend}
  (\bibinfo{year}{2015}): \emph{\bibinfo{title}{Open {S}ystem {C}ategorical
  {Q}uantum {S}emantics in {N}atural {L}anguage {P}rocessing}}.
\newblock In: {\sl \bibinfo{booktitle}{Proceedings of the 6th {C}onference on
  {A}lgebra and {C}oalgebra in {C}omputer {S}cience (CALCO)}},
  \bibinfo{address}{Nijmegen, Netherlands},
  \doi{10.4230/LIPIcs.CALCO.2015.270}.

\bibitemdeclare{article}{FrobMeanI}
\bibitem{FrobMeanI}
\bibinfo{author}{M.~\surnamestart Sadrzadeh\surnameend},
  \bibinfo{author}{S.~\surnamestart Clark\surnameend} \&
  \bibinfo{author}{B.~\surnamestart Coecke\surnameend} (\bibinfo{year}{2013}):
  \emph{\bibinfo{title}{The {F}robenius anatomy of word meanings {I}: subject
  and object relative pronouns}}.
\newblock {\sl \bibinfo{journal}{Journal of Logic and Computation}}
  \bibinfo{volume}{Advance Access}, \doi{10.1093/logcom/ext044}.

\bibitemdeclare{article}{Schuetze}
\bibitem{Schuetze}
\bibinfo{author}{H.~\surnamestart Sch{\"u}tze\surnameend}
  (\bibinfo{year}{1998}): \emph{\bibinfo{title}{Automatic word sense
  discrimination}}.
\newblock {\sl \bibinfo{journal}{Computational linguistics}}
  \bibinfo{volume}{24}(\bibinfo{number}{1}), pp. \bibinfo{pages}{97--123}.

\bibitemdeclare{article}{SelingerCPM}
\bibitem{SelingerCPM}
\bibinfo{author}{P.~\surnamestart Selinger\surnameend} (\bibinfo{year}{2007}):
  \emph{\bibinfo{title}{Dagger compact closed categories and completely
  positive maps}}.
\newblock {\sl \bibinfo{journal}{Electronic Notes in Theoretical Computer
  Science}} \bibinfo{volume}{170}, pp. \bibinfo{pages}{139--163},
  \doi{10.1016/j.entcs.2006.12.018}.

\bibitemdeclare{incollection}{SelingerSurvey}
\bibitem{SelingerSurvey}
\bibinfo{author}{P.~\surnamestart Selinger\surnameend} (\bibinfo{year}{2011}):
  \emph{\bibinfo{title}{A survey of graphical languages for monoidal
  categories}}.
\newblock In \bibinfo{editor}{B.~\surnamestart Coecke\surnameend}, editor: {\sl
  \bibinfo{booktitle}{New Structures for Physics}}, \bibinfo{series}{Lecture
  Notes in Physics}, \bibinfo{publisher}{Springer-Verlag}, pp.
  \bibinfo{pages}{275--337}, \doi{10.1007/978-3-642-12821-9}.

\bibitemdeclare{unpublished}{vandeWeteringen}
\bibitem{vandeWeteringen}
\bibinfo{author}{J.~M.~M. \surnamestart van~de Weteringen\surnameend}:
  \emph{\bibinfo{title}{A Classification of Entropic Partial Orders}}.
\newblock \bibinfo{note}{Preprint}.

\end{thebibliography}

\end{document}